\documentclass{article}

\PassOptionsToPackage{numbers, compress}{natbib}
\usepackage[main,final]{neurips_2026}

\usepackage[utf8]{inputenc}
\usepackage[T1]{fontenc}
\usepackage{hyperref}
\usepackage{url}
\usepackage{booktabs}
\usepackage{amsfonts}
\usepackage{nicefrac}
\usepackage{microtype}
\usepackage{xcolor}

\definecolor{rowblue}{RGB}{232, 240, 254}  % soft highlight for proposed rows

\usepackage{bm}
\usepackage{amsmath}
\usepackage{amssymb}
\usepackage{amsthm}
\usepackage[english]{babel}
\usepackage{multirow}
\usepackage{graphicx}
\usepackage{arydshln}
\usepackage{caption}
\usepackage{subcaption}
\usepackage{algorithm}
\usepackage{algpseudocode}
\usepackage{pifont}
\usepackage{colortbl}
\usepackage{tikz}
\usepackage{xspace}

\usetikzlibrary{fit,positioning,backgrounds,calc,quotes}
\def\model{ReinPatch\xspace}

\newcommand{\customfootnotetext}[2]{{%
\renewcommand{\thefootnote}{#1}%
\footnotetext[0]{#2}}}

\newcommand{\maybeprintauthornotes}{}
\makeatletter
\renewcommand{\maybeprintauthornotes}{%
  \if@anonymous\else
    \customfootnotetext{}{$\dag$ Correspondence to: <\texttt{yulun.wu@capitalone.com, sravan.ankireddy@utexas.edu}>}
    \customfootnotetext{}{$\P$ Capital One, \; $\S$ University of Texas, Austin}
    \customfootnotetext{}{* Equal Contribution}
  \fi
}
\makeatother

\mathchardef\mhyphen="2D

\title{Dynamic Tokenization via Reinforcement Patching: End-to-end Training and Zero-shot Transfer}
\author{
  Yulun Wu \textsuperscript{$\dag$}\;\textsuperscript{$\P$}\;\textsuperscript{*}
  \And
  Sravan Kumar Ankireddy \textsuperscript{$\dag$}\;\textsuperscript{$\S$}\;\textsuperscript{*}
  \And
  Samuel Sharpe \textsuperscript{$\P$}
  \AND
  Nikita Seleznev \textsuperscript{$\P$}
  \And
  Dehao Yuan \textsuperscript{$\P$}
  \And
  Hyeji Kim \textsuperscript{$\S$}
  \And
  Nam H. Nguyen \textsuperscript{$\P$}
}

\begin{document}

\maketitle
\maybeprintauthornotes

\begin{abstract}
Efficiently aggregating spatial or temporal horizons to acquire compact representations has become a unifying principle in modern deep learning models, yet learning data-adaptive representations for long-horizon sequence data, especially continuous sequences like time series, remains an open challenge. While fixed-size patching has improved scalability and performance, discovering variable-sized, data-driven patches end-to-end often forces models to rely on soft discretization, specific backbones, or heuristic rules. In this work, we propose Reinforcement Patching (\model), a novel framework that jointly optimizes a sequence patching policy and its downstream sequence backbone model with reinforcement learning. By formulating patch boundary placement as a discrete decision process optimized via Group Relative Policy Gradient (GRPG), \model bypasses the need for continuous relaxations and performs dynamic patching policy optimization in a natural manner% compatible with both contextual and causal prediction tasks
. Moreover, our method allows strict enforcement of a desired compression rate, freeing the downstream backbone to scale efficiently, and natively supports multi-level hierarchical modeling. We evaluate \model on time-series forecasting datasets, where it demonstrates compelling performance compared to state-of-the-art data-driven patching strategies. Furthermore, our detached design allows the patching module to be extracted as a standalone foundation patcher, providing the community with visual and empirical insights into the segmentation behaviors preferred by a purely performance-driven neural patching strategy.
\end{abstract}

\section{Introduction}
\label{sec:introduction}

A unifying principle in modern deep learning across domains is the efficient transformation of raw signals into rich representations, compressing spatial or temporal horizons into hidden dimensions through either architectural design or strategic pre-processing. In language modeling, algorithms such as byte-pair encoding (BPE)~\cite{sennrich2016neural} have become a standard pre-processing technique, condensing character-level inputs into compact, semantically meaningful sub-words that are subsequently projected into continuous embedding spaces. Similarly, in computer vision, segmenting an image into non-overlapping patches and projecting each patch independently has emerged as a canonical architectural choice, forming the basis of powerful Vision Transformers (ViT)~\cite{dosovitskiy2021image}. The shared success of these tokenization strategies highlights a broader principle: how an input signal is represented is often as important as the choice of backbone architecture itself.

Although fixed-size patching and downsampling within convolutional neural network~\cite{lecun2015deep} blocks are standard strategies for embedding compression in vision models, the same does not always hold true for sequence data such as language and time series. Instead, the inductive bias of variable-sized, data-adaptive segmentation strategies may often align better with the inherent nature of these sequences. For instance, the most prevalent tokenization approach in sequence applications such as language has been compression via pre-processing (i.e. BPE tokenization) instead of via neural modeling. However, algorithms designed for discrete sequence tokenization such as BPE are not naturally well-suited for other sequences such as time series due to the continuous nature of these data. Therefore, many early transformer-based approaches in time series~\cite{zhou2021informer, wu2021autoformer, zhou2022fedformer} did not have a compression component and projected each individual time step independently into the embedding space, producing prohibitively long sequences that scale poorly because of the quadratic complexity of self-attention. More recently, PatchTST~\cite{nie2023patch} experimented with fixed-size, non-data-adaptive patching on time series, introducing a simple but effective mechanism that groups a fixed number of contiguous time steps into a \textit{patch} and represents each patch with a single embedding, thereby compressing the input signal and shortening the sequence length. Since then, some recent work have explored data-driven strategies for selecting patch sizes or boundary locations to replace static patching, either by combining existing patches to form larger patches and richer representations~\cite{huang2024hdmixer,wuenhancing} or by relying on auxiliary external rules~\cite{ankireddytimesqueeze, wang2025lightgts} or external models~\cite{abeywickrama2025entrope}. Even for language models, more recent works such as BLT~\cite{pagnoni2025byte} and H-Net~\cite{hwang2025dynamic} have also investigated using various patching strategies (also known as \textit{chunking}) as part of the neural model design to replace BPE tokenizer.

While much of mainstream deep learning research has focused on language and vision, time-series data remains central to many high-impact domains, including energy, finance, climate, and healthcare. Furthermore, it is a domain mostly consisting of long-horizon continuous data where patching strategies are crucial to methodology design. The simple static patching approach in PatchTST alone led to strong forecasting gains, demonstrating that efficient compression is a critical factor even when the backbone architecture remains unchanged.

\begin{table*}[t]
\centering
\small
% Optional for symbols/colors: \usepackage{pifont,xcolor,colortbl}
\providecommand{\cmark}{\textcolor{green!60!black}{\ding{51}}}
\providecommand{\xmark}{\textcolor{red!75!black}{\ding{55}}}
\providecommand{\pmark}{\textcolor{orange!90!black}{\ding{51}/\ding{55}}}
\providecommand{\na}{n/a}
\caption{Qualitative comparison of \model with existing patching strategies across key design properties. \textbf{Characteristics:} \textbf{Adaptive} (Data-adaptive); \textbf{Var.\ Size} (Variable-sized); \textbf{Model} (Model-based); \textbf{Heuristic-free} (Unreliant on human-designed heuristics or rules); \textbf{Agnostic} (Unrestricted to particular data types); \textbf{Hierarchical} (Supports hierarchical patching). \textbf{Optimization:} \textbf{E2E} (End-to-end); \textbf{Non-local} (Unlimited to local exploration during end-to-end training); \textbf{Detached} (Detached from the downstream sequence model). \cmark: satisfied, \xmark: not satisfied, \pmark: partially satisfied, \na: not applicable.}
\label{tab:patch_comparison}
    
\setlength{\tabcolsep}{4.5pt}
\renewcommand{\arraystretch}{1.12}

\resizebox{\textwidth}{!}{%
\begin{tabular}{lcccccc ccc}
\toprule
& \multicolumn{6}{c}{\textbf{Characteristics}} & \multicolumn{3}{c}{\textbf{Optimization}} \\
\cmidrule(lr){2-7} \cmidrule(lr){8-10}
\textbf{Method} & 
\textbf{Adaptive} & 
\textbf{Var. Size} & 
\textbf{Model} & 
\textbf{Heuristic-free} & 
\textbf{Agnostic} & 
\textbf{Hierarchical} & 
\textbf{E2E} & 
\textbf{Non-local} & 
\textbf{Detached} \\
\midrule
Static Patching~\cite{nie2023patch} & \xmark & \xmark & \xmark & \cmark & \cmark & \cmark & \xmark & \na & \cmark \\
Entropy Patching~\cite{pagnoni2025byte} & \cmark & \cmark & \cmark & \xmark & \xmark & \xmark & \xmark & \na & \cmark \\
TimeSqueeze's Compression~\cite{ankireddytimesqueeze} & \cmark & \cmark & \xmark & \xmark & \xmark & \xmark & \xmark & \na & \cmark \\
H-Net's Chunking~\cite{hwang2025dynamic} & \cmark & \cmark & \cmark & \xmark & \cmark & \cmark & \cmark & \xmark & \xmark \\
Periodic Patching~\cite{wang2025lightgts} & \cmark & \xmark & \xmark & \xmark & \xmark & \xmark & \xmark & \na & \cmark \\
Selective Patching~\cite{wuenhancing} & \cmark & \xmark & \cmark & \xmark & \cmark & \xmark & \pmark & \xmark & \xmark \\
\rowcolor{gray!10}
\model (Ours) & \cmark & \cmark & \cmark & \cmark & \cmark & \cmark & \cmark & \cmark & \cmark \\
\bottomrule
\end{tabular}%
}

\end{table*}

In this work, we further explore the end-to-end optimization of patching strategy alongside the training of deep sequence models. In contrast to prior works such as H-Net~\cite{hwang2025dynamic} and SRSNet~\cite{wuenhancing}, which incorporate a patching component into the neural design of their main backbone model and rely on continuous relaxations of patching boundary placement, we look at this problem through a different angle and use reinforcement learning to naturally and canonically solve what is essentially a discrete action optimization problem. This means we need not rely on soft discretization of patching boundaries to enable gradient flow, and we allow the patching policy to be independent of the neural design of the backbone model. We elaborate the benefits of doing so throughout Section~\ref{sec:related-work} and Section~\ref{sec:proposed-method}, and provide a summary in advance in Table~\ref{tab:patch_comparison} to give the reader an overview of the advantages of our approach. Our contributions are as follows:
\begin{itemize}
    \item We propose \textit{Reinforcement Patching}, or \textit{\model} for short, which to our knowledge is the first framework to use reinforcement learning for joint optimization of a deep sequence model and a sequence patching policy in an end-to-end fashion while keeping the patching policy as a separate neural module and treating the deep sequence model as its \textit{environment}.
    \item This separation allows a standalone foundation patcher to be established after large-scale joint pre-training, and allows continued pre-training of the deep sequence model with the patcher's weights frozen. As a proof of concept, we produce a foundation patcher on univariate time series and benchmark its zero-shot performance in our experiments.
    \item We focus on time-series forecasting in experiments and show that our method improves over a broad set of fixed- and variable-size patching strategies under a unified hierarchical backbone on multiple forecasting datasets. We perform analysis and ablation studies to help readers understand the behavior of \model, and visualize the patches it generates to provide insight into the segmentation patterns preferred by a performance-driven neural patching model.
\end{itemize}

\section{Related Work}
\label{sec:related-work}

Recent advancements in sequence modeling have heavily relied on static patching methods, popularized by architectures like PatchTST~\citep{nie2023patch}, which use uniform patch sizes to improve training and inference efficiency. This design has been adopted by various time-series foundation models, including TimesFM~\citep{das2024decoder}, MOMENT~\citep{goswami2024moment}, Moirai~\citep{woo2024unified}, and Timer-XL~\citep{liu2024timer}. However, static patching struggles to capture substantial temporal heterogeneity within real-world series. To address this, several data-adaptive strategies have emerged, such as HDMixer~\citep{huang2024hdmixer}, LightGTS~\citep{wang2025lightgts}, and SRSNet~\citep{wuenhancing}, which automatically adjust patch sizes or combine fixed-size patches. Yet, these methods do not explicitly learn dynamic, within-sequence policies based on local signal characteristics.

Other approaches rely on external rules or auxiliary models to determine boundaries, such as BLT and its variants~\citep{pagnoni2025byte,abeywickrama2025entrope,katariya2026dp}, which use a separately trained entropy model, SpaceByte~\citep{slagle2024spacebyte}, which uses natural word separators, or TimeSqueeze~\citep{ankireddytimesqueeze}, which uses local variance. Because these depend on modality-specific heuristics or external identifiers, they cannot be jointly optimized with the main sequence backbone. Alternatively, end-to-end patching methods, like H-Net~\citep{hwang2025dynamic}, enable gradient-based patch learning through smoothed routing, but this tightly couples the patching strategy to the main model's training. This coupling introduces several limitations, such as preventing a regime where the patching module is pre-trained and frozen, and restricting the approach strictly to causal modeling due to the architectural design of the main model. Besides, H-Net has a few other drawbacks compared to our proposed method in terms of policy exploration, model pre-planning, and inductive bias heuristics. We will further elaborate on these topics in the following section. A more extended discussion of related work can be found in Appendix~\ref{appendix:extended_related_work}.

\section{Proposed Method}
\label{sec:proposed-method}

\begin{figure*}%[hbt!]
    \centering
        \includegraphics[width=0.9\textwidth]{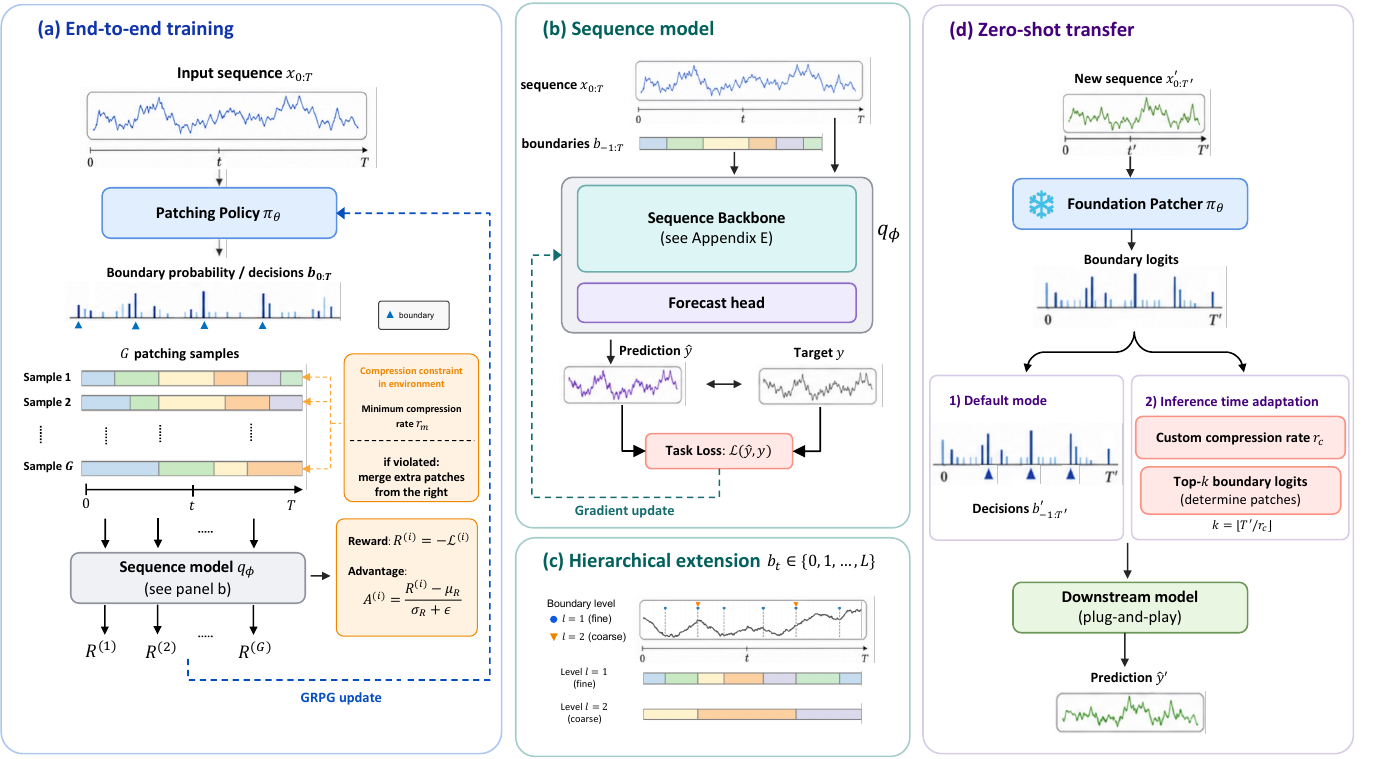}
        \caption{\model learns a patching \textbf{policy $\pi_\theta$} decoupled from the main sequence \textbf{backbone $q_\phi$}. \textbf{(a)} \textit{Training}: for each input sequence, the policy $\pi_\theta$ emits per-position boundary logits. We sample $G$ boundary sets, each representing a patching strategy of the sequence that the backbone $q_\phi$ consumes to produce a prediction. The negative task loss on this prediction is then the terminal reward for the corresponding patching strategy. Advantages are computed group-relative and $\pi_\theta$ is updated by GRPG. \textbf{(b)} \textit{Sequence model}: encoder downsamples points into patches, a latent transformer models patches, a decoder upsamples patches back to base granularity (Appendix~\ref{appendix:model_architecture}). A forecast head then generates the prediction to calculate the task loss. Backbone $q_\phi$ is updated through standard loss back-propagation. \textbf{(c)} \textit{Hierarchical}: the same policy can be extended to emit multi-level boundaries in one shot, without having to plan on-the-fly using the hidden representations of $q_\phi$ throughout its forward pass. \textbf{(d)} \textit{Zero-shot transfer}: after large-scale pre-training, learned patching policy $\pi_\theta$ can be frozen and used as a standalone module to produce boundaries for a new downstream backbone, optionally at a custom compression rate via top-$k$ on the boundary logits.}
    \label{fig:framework_overview}
\end{figure*}

In this section, we describe the formulation, optimization, and adaptation of our proposed method in details. For an overview of the method, see Figure~\ref{fig:framework_overview}.

\subsection{Modeling Framework}

Let $\mathbf{x}_T = x_{0:T}$ be a sequence data with initial point $x_0$ and $T$ subsequent evolutions $\{x_1, \dots, x_T\}$. Let $\mathbf{b}_T = b_{-1:T}$ be an auxiliary sequence of $\mathbf{x}_T$ where $b_t \in \{0, 1\}$ for $0 \leq t \leq T$ is a right-side boundary decision for data point $x_t$ (and $b_{-1} = 1$). Our goal is to find the best $b_{0:T}$ such that a deep sequence model $q_\phi(\mathbf{x}_T, \mathbf{b}_T)$ optimizes its objective $\mathcal{L}$ under the patching strategy induced by $\mathbf{b}_T$. For example, if $T = 5$ and $\mathbf{b}_T = \{1,0,0,1,0,1,1\}$, then $q_\phi$ will patch the input sequence according to the following partition: $\mid x_0, x_1, x_2 \mid x_3, x_4 \mid x_5 \mid$ during its embedding stage. Note that the last boundary is relevant in the generative setting, where the objective is $\sum_{t=0}^{T-1}\mathcal{L}(q_\phi(x_{0:t}, b_{-1:t}), x_{t+1})$ and the next point $x_{T+1}$ is generated through $q_\phi(\mathbf{x}_T, \mathbf{b}_T)$ during inference time (see Appendix~\ref{appendix:extended_modeling_framework} for details). The patching policy $\pi_\theta$ that generates boundaries $b_{0:T}$ is then optimized according to the performance of sequence model $q_\theta$ w.r.t. $\mathcal{L}$.

For sequence modeling tasks that do not abide by strict time causality during representation learning, such as time series forecasting, masked language modeling, and sequence classifications, the full contextual sequence from initial to terminal state is presented to the model at once. Therefore, the policy network $\pi_{\theta}$ to decide $b_{0:T}$ can make a joint, collective decision in one step regarding which places to set the patching boundaries, effectively transforming the formulation into a contextual bandit problem. To concretize, we define state $s \triangleq \mathbf{x}_T$ and action $a \triangleq \mathbf{b}_T$, and subsequently formulate the policy and reward as:
\begin{align}
    \label{eq:contextual-bandit}
    \pi(a \mid s) \triangleq \pi_\theta \left(b_{0:T} \mid \mathbf{x}_T \right) \cdot \delta_{b_{-1}}, \quad R(s, a) \triangleq -\mathcal{L} \left(q_\theta \left(\mathbf{x}_T, \mathbf{b}_T \right), y \right)
\end{align}
where $y$ is the prediction target of the given task, $\delta_k$ is the Dirac delta distribution with nonzero density only at $k$. Under this setup, the reward function depends solely on the sequence model $q_\phi$ and its canonical objective $\mathcal{L}$, setting the stage for a simple and efficient joint optimization scheme for both $\pi_\theta$ and $q_\phi$ as we will elaborate in the next section. While our main experiments focus on this non-causal contextual setting, ReinPatch natively supports generative causal modeling via a multi-step MDP formulation. We provide the full unified multi-step MDP formulation in Appendix~\ref{appendix:extended_modeling_framework}.

\subsection{Objectives and Optimization}

Under this formulation, the simultaneous optimization of $q_\phi$ and $\pi_\theta$ is clean and straightforward. Since the transition dynamics $P$ and reward function $R$ are only dependent on the main backbone sequence model $q_\theta$ and its objective $\mathcal{L}$, the forward pass of $q_\theta$ can essentially be seen as the gym environment for $\pi_\theta$. Hence, after each standard batch gradient update of the sequence model $q_\theta$, its losses can be directly used to compute $\pi_\theta$'s objective. Notice that, similar to \citet{shao2024deepseekmath}, we frame the full sequence of actions as a whole in optimization for efficiency, such that they are judged by the same terminal reward and same sequence-level critic, despite having a general MDP formulation (Appendix~\ref{appendix:extended_modeling_framework}). This is essentially what makes group-relative approaches viable and desirable. We use a simple group relative policy gradient algorithm as the objective of $\pi_\theta$:
\begin{align}
    \label{eq:grpg}
    %\nabla_\theta \mathcal{J}_{GRPG} = \frac{1}{G} \sum_{i=1}^G \mathbb E \left[ \sum_{t=0}^T \nabla_\theta \log \pi_\theta (a_t^{(i)} \mid s_t^{(i)}) \frac{R(s_T^{(i)}, a_T^{(i)}) - \mu_R}{\sigma_R + \epsilon}\right]
    \nabla_\theta \mathcal{J}_{GRPG} = \frac{1}{G} \sum_{i=1}^G \mathbb E \left[ \nabla_\theta \log \pi_\theta (a^{(i)} \mid s^{(i)}) \frac{R(s_T^{(i)}, a_T^{(i)}) - \mu_R}{\sigma_R + \epsilon}\right]
\end{align}
where $\mu_R$ and $\sigma_R$ are the mean and standard deviation of $G$ terminal rewards $\{R(s_T^{(i)}, a_T^{(i)})\}_{1 \leq i \leq G}$, since training an additional critic network as baseline~\cite{mnih2016asynchronous} would often be an overkill for terminal-reward-only settings, let along one-step MDP settings. For reasons why GRPO~\citep{shao2024deepseekmath} is not preferred over GRPG under our setup, see Appendix~\ref{appendix:further_modeling_discussion}.

To put it in words, we sample $G$ different sets of boundaries from policy $\pi_\theta$ for each input sequence, then decide their relative advantage based on how well the sequence model $q_\phi$ performs on this sequence under each set. Note that standardization can be replaced by centering (i.e. replace $\sigma_R + \epsilon$ with $1$) depending on the data characteristics, as the latter can preserve the importance of decisions. %For example, if the choice of patching boundaries matters enormously for some sequences in the dataset while remains trivial for the others, the policy would not be forced to learn intensely over the latter under the same magnitude of the former. 
By solving patching boundary optimization with RL, we do not rely on soft-discretization to enable gradient flow like prior work such as H-Net~\citep{hwang2025dynamic}, which entails that the policy exploration space is continuous, only residing in the local neighborhood of the current boundary policy.

\paragraph{Enforcing desired compression rate} To prevent specification gaming, our approach sets a strict minimum compression rate $r_m$ in the \textit{environment} rather than using a soft auxiliary loss as in \citet{hwang2025dynamic}. If $\pi_\theta$ proposes boundaries yielding a lower compression rate than $r_m$, extra patches on the right side will be merged into one during downsampling. See Algorithm~\ref{alg:reinpatch_training} in Appendix~\ref{appendix:training_algorithm} for details. This strict enforcement forces the policy to adapt to the constraint in the environment naturally during training, and more importantly, frees practitioners from pre-planning model sizes for the worst-case (uncompressed) scenario and allowing the model to reap the full scaling benefits of patching. We elaborate on the systematic advantages of this strict enforcement in Appendix~\ref{appendix:further_modeling_discussion}.

\subsection{Backbone Architecture}
\label{sec:backbone-arch}

For patching policy $\pi_\theta$, we design it as a standalone lightweight transformer such that it can be easily extracted as a foundation patching policy under the joint pre-training setup, to use for downstream external chunking on smaller datasets. It is composed of an embedding model $m_\theta: \mathbb R^d \to \mathbb R^{d_{patch}}$ and a final linear layer $f_\theta: \mathbb R^{d_{patch}} \to \mathbb R^2$ to project the embedding into the corresponding logits for decisions $\{ 0, 1 \}$. In contrast, prior end-to-end patching model such as H-Net~\cite{hwang2025dynamic} has far more complicated patching mechanisms, which rely on the human perception that it is best to set boundaries when context shifts, just like prior external-rule-based and external-model-based methods. This may not necessarily be true for the optimal patching strategy learned by a performance-driven model.

For downstream sequence model $q_\phi$, we use a clean and unified architecture for the ease of benchmarking and ablation, demonstrated in Figure~\ref{fig:model_arch} of Appendix~\ref{appendix:model_architecture}. The architecture allows downsampling and upsampling to be done either by cross-attention encoder and decoder, as in \citet{pagnoni2025byte}, or simply scatter and gather plus residual, as in \citet{hwang2025dynamic}.

\paragraph{Extension to multi-level hierarchical model.} The extension to hierarchical model (i.e. more than one level of patching) is elegant under our proposed method, since the patching policy net need not necessarily observe the evolving hidden representation under its downstream sequence model $q_\phi$, and therefore is able to plan multi-level patching strategy at once. For sequence model $q_\phi$, the extension to hierarchical is trivial (see Appendix~\ref{appendix:model_architecture}); for patching policy $\pi_\theta$, we apply a simple modification to final linear layer $f_\theta$ such that $f_\theta: \mathbb R^{d_{patch}} \to \mathbb R^L$ outputs the corresponding logits for extended decision set $b_t \in \{ 0, 1, \dots, L \}$ if $L$ levels of patching is desired, where $b_t = l$ means the right end of $x_t$ is the boundary for the first $l$ levels of patching. This is well-defined since boundaries on the $j$-th level will always be a subset of boundaries on the $i$-th level, for any $i < j$.

\subsection{Inference Time Adaptation}
\label{sec:inference-adaptation}

In order to give the foundation patcher additional flexibility to scale across various downstream tasks, where the downstream sequence model width and dataset size could be significantly different from the ones in its pre-training, it should be able to adapt its patching aggressiveness according to a custom compression rate that the practitioners deem more suited to their downstream settings. Hence, in inference mode, we provide the learned patching policy $\pi_\theta$ with the ability to take a target compression rate $r_c$ and adapt accordingly. For contextual tasks, this is relatively simple: we compute the boundary logits $\{l_t\}_{0 \leq t \leq T}$ with $\pi_\theta$ and pick the indices corresponding to the top-$k$ logits where $k = \lfloor T / r_c \rfloor$. If a target compression rate is not specified by the practitioner, the patching policy calculates the empirical expectation of $\hat{k} = \sum_{t=0}^T \left( 1 + e^{-l_t}\right)^{-1}$ and pick indices by the top-$\lfloor \hat{k} \rfloor$ logits instead. For generative tasks (i.e. causal tasks), this is more challenging because it is an \textit{online thresholding problem} where we need to hit a global budget (the target compression rate) without seeing the future distribution of the boundary probabilities. The classic approach is to keep a sliding window of logits (or exponential moving mean and variance) and place patching boundary only if the current logit is higher than the $(1-1/r_c)$-th quantile. Note that the yielded compression rate in this case does not strictly adhere to the target compression rate as it does in the contextual tasks, and there are pathological cases that could break this approach (for example, a monotone increasing set of logits, which results in boundary placement everywhere). To this end, we leave further thoughts and development regarding inference time adaptation for generative tasks to future research.

\section{Experiments}
\label{sec:experiments}

\subsection{Experimental Details}

\textbf{Task.}
We evaluate \model on multivariate time series forecasting tasks (problem formulation in Appendix~\ref{appendix:problem-formulation}), where the appropriate patching strategy depends both on dataset-level characteristics and on local temporal dynamics. This makes long-term forecasting a suitable setting for comparing different patching strategies. Following prior work~\cite{zhou2021informer,wu2021autoformer}, we evaluate on the four ETT subsets together with Weather, Electricity, Solar, and Traffic. %These benchmarks cover multiple domains, sampling frequencies, dimensionalities, and train-validation-test splits, and therefore provide a diverse testbed for evaluating sequence patching strategies.
Details of the datasets used in our forecasting evaluation are presented in Table~\ref{tab:dataset_stats} of Appendix~\ref{appendix:experiment-details}.

\textbf{Backbone.}
To evaluate whether \model improves forecasting performance relative to existing fixed-size and variable-size patching strategies, we adopt a generic model architecture that has been widely used in patch-based forecasting models as described in Appendix \ref{appendix:model_architecture}. Across all methods, we use this model architecture and vary only the patching algorithm as long as its associated encoder-decoder configuration. This controlled protocol isolates the effect of the patching strategy and enables a fair comparison. Additional architecture and selection details are provided in Appendix~\ref{appendix:experiment-details}.

% Once the contextualized full-resolution embeddings are acquired, direct multi-step forecasting is performed where a simple linear head is used to directly predict the full forecasting horizon from the flattened embeddings~\cite{nie2023patch, zeng2023transformers}.
%Specifically, let $\mathbf{Z} \in \mathbb{R}^{N \times d}$ denote the output token embeddings from the decoder, where $N$ is the total number of output tokens and $d$ is the model's hidden dimension. The embeddings are flattened into a single vector $\mathbf{z} = \operatorname{Flatten}(\mathbf{Z}) \in \mathbb{R}^{N \cdot d}$, which is then projected to the target forecasting horizon $H$ via a single linear transformation: $\hat{\mathbf{y}} = \mathbf{W}\,\mathbf{z} + \mathbf{b}$, where $\mathbf{W} \in \mathbb{R}^{H \times (N \cdot d)}$ and $\mathbf{b} \in \mathbb{R}^{H}$ are learnable parameters, and $\hat{\mathbf{y}} \in \mathbb{R}^{H}$ is the predicted output sequence. The entire model is trained end-to-end by minimizing the mean squared error between the predictions $\hat{\mathbf{y}}$ and the ground-truth future values $\mathbf{y} \in \mathbb{R}^{H}$: $\mathcal{L} =  \left\| \hat{\mathbf{y}} - \mathbf{y} \right\|_2^2$.

\textbf{Baselines.}
We benchmark \model against an extensive collection of fixed- and variable-size patching baselines. On the fixed-size side, static patching uses a predetermined uniform patch size~\cite{nie2023patch}, selective patching constructs additional features from overlapping fixed-size patches using a scorer~\cite{wuenhancing}, and periodic patching chooses the patch size based on signal periodicity in the Fourier domain~\cite{wang2025lightgts}. On the variable-size side, entropy patching relies on entropy estimation from an external pretrained model~\cite{pagnoni2025byte,abeywickrama2025entrope}, TimeSqueeze compression uses local variance in the raw input space~\cite{ankireddytimesqueeze}, and H-Net chunking learns patch boundaries end to end from cosine similarity between Mamba embeddings computed at full resolution~\cite{hwang2025dynamic}.

\subsection{Main Results}
\label{sec:main_results}
Following standard practice, we report Mean Squared Error (MSE) and Mean Absolute Error (MAE) across four forecasting horizons, $H \in \{96, 192, 336, 720\}$. Because the look-back window size heavily influences performance, we evaluate all methods across look-back lengths $T \in \{96, 192, 336\}$ and report the best validation-selected configuration for each. Additional training details can be found in Appendix~\ref{appendix:training-selection-details}.% Red and blue table entries indicate first and second place.
% according to the unrounded values used for ranking; ties at the displayed precision should be read as comparable.

\textbf{End-to-end patching.} We first evaluate \model in an end-to-end setting, jointly learning the patching policy and the forecasting model on each dataset. Table~\ref{tab:results_avg} presents the best average MSE and MAE across all forecasting horizons. \model ranks first or second in average MSE on six of the eight datasets, and it gives the best average MSE on ETTh2, ETTm1, and Traffic. The baseline values are averaged over the same five-seed protocol, with baseline standard deviations omitted from the table for readability. These consistent improvements under the shared forecasting backbone indicate that the performance gains stem from the learned, data-driven patching policy rather than architectural modifications.

\textbf{Zero-shot patching.} A broader objective of our work is to explore the possibility of establishing a foundation time series patcher by pre-training a policy on a large-scale corpus for off-the-shelf, zero-shot application on smaller target datasets. To achieve this, we jointly pre-train the patcher and a main sequence forecasting model on the Unified Time Series Dataset (UTSD)~\cite{liutimer}. UTSD aggregates diverse, multi-frequency data from public repositories and real-world operations across seven domains: Energy, Environment, Health, IoT, Nature, Transportation, and Web. For downstream evaluation, we deploy this foundation patcher in a zero-shot configuration by freezing its pre-trained weights and exclusively training the task-specific forecasting backbone.

\begin{table*}
\caption{Multivariate forecasting average results with forecasting horizons $H \in \{96, 192, 336, 720\}$ for the datasets. \textcolor{red}{\textbf{Red}}: the best, \textcolor{blue}{\underline{Blue}}: the 2nd best. Full results are presented in Table~\ref{tab:results_full} of Appendix~\ref{appendix:additional_results}.}
\label{tab:results_avg}
\centering
\fontsize{7}{8}\selectfont
\renewcommand{\arraystretch}{1.0}
\setlength{\tabcolsep}{4pt}
\resizebox{\textwidth}{!}{
    \begin{tabular}{@{}lcccccccccccccccc@{}}
    \toprule
        \textbf{Datasets} & \multicolumn{2}{c}{\textbf{ETTh1}} & \multicolumn{2}{c}{\textbf{ETTh2}} & \multicolumn{2}{c}{\textbf{ETTm1}} & \multicolumn{2}{c}{\textbf{ETTm2}} & \multicolumn{2}{c}{\textbf{Weather}} & \multicolumn{2}{c}{\textbf{Electricity}} & \multicolumn{2}{c}{\textbf{Solar}} & \multicolumn{2}{c}{\textbf{Traffic}} \\
        \cmidrule(lr){2-3} \cmidrule(lr){4-5} \cmidrule(lr){6-7} \cmidrule(lr){8-9} \cmidrule(lr){10-11} \cmidrule(lr){12-13} \cmidrule(lr){14-15} \cmidrule(lr){16-17}
        \textbf{Metrics} & \textbf{MSE} & \textbf{MAE} & \textbf{MSE} & \textbf{MAE} & \textbf{MSE} & \textbf{MAE} & \textbf{MSE} & \textbf{MAE} & \textbf{MSE} & \textbf{MAE} & \textbf{MSE} & \textbf{MAE} & \textbf{MSE} & \textbf{MAE} & \textbf{MSE} & \textbf{MAE} \\
        \midrule
        Static Patching~\cite{nie2023patch} & 0.409 & 0.430 & 0.335 & 0.383 & 0.352 & 0.385 & 0.272 & 0.326 & 0.230 & 0.267 & 0.194 & 0.298 & 0.208 & 0.267 & 0.431 & 0.304 \\
        Selective Patching~\cite{wuenhancing} & 0.408 & 0.430 & 0.331 & 0.383 & 0.359 & 0.392 & 0.268 & 0.324 & 0.230 & 0.269 & 0.187 & 0.292 & 0.221 & 0.278 & 0.453 & \textcolor{red}{\textbf{0.286}} \\
        Periodic Patching~\cite{wang2025lightgts} & 0.413 & 0.433 & 0.336 & 0.384 & 0.356 & 0.388 & 0.272 & 0.326 & \textcolor{blue}{\underline{0.228}} & 0.267 & 0.191 & 0.294 & 0.210 & 0.266 & 0.449 & 0.287 \\
        Entropy Patching~\cite{pagnoni2025byte,abeywickrama2025entrope} & 0.425 & 0.438 & 0.339 & 0.389 & 0.358 & 0.386 & 0.281 & 0.334 & 0.230 & 0.267 & \textcolor{blue}{\underline{0.169}} & \textcolor{red}{\textbf{0.267}} & 0.203 & \textcolor{red}{\textbf{0.261}} & 0.480 & 0.338 \\
        TimeSqueeze's Compression~\cite{ankireddytimesqueeze} & 0.419 & 0.433 & 0.334 & 0.385 & 0.362 & 0.387 & 0.271 & 0.327 & 0.229 & 0.269 & 0.175 & 0.275 & 0.204 & 0.269 & 0.502 & 0.318 \\
        H-Net's Chunking~\cite{hwang2025dynamic} & 0.434 & 0.437 & 0.339 & 0.387 & 0.362 & 0.384 & 0.272 & 0.327 & 0.229 & \textcolor{red}{\textbf{0.265}} & 0.175 & 0.275 & 0.209 & 0.275 & 0.693 & 0.419 \\
        \midrule
        \model\ (end-to-end) [Ours] & \textcolor{blue}{\underline{0.405}} & \textcolor{blue}{\underline{0.428}} & \textcolor{red}{\textbf{0.323}} & \textcolor{blue}{\underline{0.378}} & \textcolor{red}{\textbf{0.347}} & \textcolor{red}{\textbf{0.382}} & \textcolor{blue}{\underline{0.263}} & \textcolor{red}{\textbf{0.321}} & 0.229 & 0.268 & 0.172 & 0.273 & \textcolor{blue}{\underline{0.202}} & 0.267 & \textcolor{red}{\textbf{0.415}} & 0.288 \\
        \model\ (zero-shot) [Ours] & \textcolor{red}{\textbf{0.397}} & \textcolor{red}{\textbf{0.424}} & \textcolor{blue}{\underline{0.323}} & \textcolor{red}{\textbf{0.377}} & \textcolor{blue}{\underline{0.347}} & \textcolor{blue}{\underline{0.382}} & \textcolor{red}{\textbf{0.262}} & \textcolor{blue}{\underline{0.321}} & \textcolor{red}{\textbf{0.226}} & \textcolor{blue}{\underline{0.265}} & \textcolor{red}{\textbf{0.168}} & \textcolor{blue}{\underline{0.267}} & \textcolor{red}{\textbf{0.199}} & \textcolor{blue}{\underline{0.263}} & \textcolor{blue}{\underline{0.416}} & \textcolor{blue}{\underline{0.286}} \\
        \bottomrule
    \end{tabular}}
\end{table*}

As shown in Table~\ref{tab:results_avg}, this zero-shot configuration generally outperforms the end-to-end training of \model. It is at least tied with the end-to-end variant on seven of eight datasets in average MSE and on all eight datasets in average MAE, yielding the best mean rank among all methods for both metrics. Its mean rank is $1.125$ for both MSE and MAE. This suggests that exposing the patching policy to a massive, highly diverse corpus yields more robust and generalizable patches than training exclusively on a narrower target dataset. By freezing the pre-trained weights, the zero-shot patcher avoids dataset-specific overfitting and effectively regularizes the downstream forecasting backbone. This transferability is also by design: the policy is optimized for the instance-normalized signal rather than domain-specific cues such as absolute scale or timestamp spacing, so the structural patterns it learns to segment generalize across domains and distributions.

Across the average results in Table~\ref{tab:results_avg}, compared to the entropy-based patching baseline, \model reduces average MSE by $4.9\%$ and average MAE by $3.3\%$ across datasets, with the largest gains on Traffic ($13.5\%$ MSE and $15.4\%$ MAE).
% the best \model variant achieves the lowest MSE on all eight datasets and the lowest or tied-lowest MAE on seven of eight datasets. Compared with the strongest non-\model baseline for each dataset, this corresponds to an average MSE reduction of $2.0\%$, with the largest gain of $3.7\%$ on Traffic. For MAE, \model matches or improves the strongest baseline on seven datasets and has an average relative gain of $0.5\%$, with Solar as the only exception. Compared with the frequently used entropy-based patching baseline, \model reduces average MSE by $4.9\%$ and average MAE by $3.3\%$ across datasets, with the largest gains on Traffic ($13.5\%$ MSE and $15.4\%$ MAE).  
The per-horizon results in Table~\ref{tab:results_full} of Appendix~\ref{appendix:additional_results} show the same trend at finer granularity. Across the 32 dataset-horizon settings, the zero-shot \model variant ranks first in 19 MSE entries and 15 MAE entries, while the end-to-end variant adds another 5 MSE and 3 MAE first-place entries. Overall, \model accounts for 42 of the 64 first-place metric entries in the full table. A Friedman-Nemenyi analysis over all 32 settings confirms that \model's rank advantage is statistically significant against every non-\model baseline on both metrics (Appendix~\ref{appendix:significance}).

\subsection{Ablation and Analysis}
\label{sec:ablation}

\textbf{Isolating the patching objective.} The central question is whether \model's gains stem from the clean, discrete RL objective itself or merely from the architecture and pre-training exposure. While the main results in Table~\ref{tab:results_avg} controlled for most of this exposure, they still allowed varying the encoder/decoder backbone to respect the original literature. In this section, we hold the entire architecture fixed and vary only how boundaries are learned, comparing \model (zero-shot frozen-UTSD and end-to-end) against three non-RL dynamic patchers: differentiable relaxations of the \emph{same} policy (a straight-through estimator, STE, and Gumbel-Softmax); H-Net chunking run with the \emph{identical} transformer encoder/decoder rather than its SSM variant; and an external entropy patcher pre-trained and frozen under the same UTSD protocol as our zero-shot patcher (Entropy-UTSD). Table~\ref{tab:controlled_ablation} reports mean $\pm$ std over five seeds across the 32 (dataset, horizon) settings. The zero-shot \model variant alone is best on $24/32$ (MSE) and $20/32$ (MAE) settings, rising to $30/32$ and $25/32$ when either \model variant counts, while the differentiable relaxations, the identical-backbone H-Net, and the UTSD-pretrained entropy patcher win only isolated cells and never lead in aggregate.

\begin{table*}[!htbp]
\caption{Controlled ablation of the \emph{patching objective} on a shared transformer backbone and
boundary parameterization. We compare \model (zero-shot frozen-UTSD and end-to-end) against three
non-RL dynamic patchers: differentiable relaxations of the \emph{same} policy (straight-through
estimator, STE; and Gumbel-Softmax), H-Net's chunking run with the \emph{identical} transformer
encoder/decoder (not the SSM variant), and a non-RL external entropy patcher pre-trained and frozen
under the same UTSD protocol as our zero-shot patcher (Entropy-UTSD). Cells are mean $\pm$ std over
5 seeds. \textcolor{red}{\textbf{Red}}: best, \textcolor{blue}{\underline{Blue}}: 2nd best, per (row, metric).}
\label{tab:controlled_ablation}
\centering
\scriptsize
\renewcommand{\arraystretch}{1.0}
\setlength{\tabcolsep}{3pt}
\resizebox{\textwidth}{!}{%
\begin{tabular}{@{}cc|cc|cc|cc|cc|cc|cc@{}}
\toprule
\multicolumn{2}{c|}{\textbf{Models}}
 & \multicolumn{2}{c|}{\textbf{\model\ (zero-shot) [Ours]}}
 & \multicolumn{2}{c|}{\textbf{\model\ (end-to-end) [Ours]}}
 & \multicolumn{2}{c|}{\textbf{STE Relaxation}}
 & \multicolumn{2}{c|}{\textbf{Gumbel-Softmax}}
 & \multicolumn{2}{c|}{\textbf{H-Net (transformer)}}
 & \multicolumn{2}{c}{\textbf{Entropy-UTSD}} \\
\multicolumn{1}{c|}{\textbf{Metric}} & \textbf{H}
 & \textbf{MSE} & \textbf{MAE} & \textbf{MSE} & \textbf{MAE} & \textbf{MSE} & \textbf{MAE}
 & \textbf{MSE} & \textbf{MAE} & \textbf{MSE} & \textbf{MAE} & \textbf{MSE} & \textbf{MAE} \\
\midrule
\multirow{4}{*}{\raisebox{3pt}{\rotatebox[origin=l]{90}{ETTh1}}}
 & 96  & \textcolor{red}{\textbf{0.360}} {\tiny$\pm$.008} & \textcolor{red}{\textbf{0.400}} {\tiny$\pm$.006} & 0.374 {\tiny$\pm$.002} & 0.405 {\tiny$\pm$.004} & 0.382 {\tiny$\pm$.007} & 0.405 {\tiny$\pm$.005} & 0.377 {\tiny$\pm$.005} & 0.403 {\tiny$\pm$.004} & \textcolor{blue}{\underline{0.374}} {\tiny$\pm$.005} & 0.403 {\tiny$\pm$.002} & 0.379 {\tiny$\pm$.001} & \textcolor{blue}{\underline{0.400}} {\tiny$\pm$.001} \\
 & 192 & \textcolor{red}{\textbf{0.394}} {\tiny$\pm$.003} & \textcolor{red}{\textbf{0.419}} {\tiny$\pm$.004} & \textcolor{blue}{\underline{0.410}} {\tiny$\pm$.011} & \textcolor{blue}{\underline{0.423}} {\tiny$\pm$.005} & 0.416 {\tiny$\pm$.012} & 0.430 {\tiny$\pm$.006} & 0.428 {\tiny$\pm$.008} & 0.434 {\tiny$\pm$.006} & 0.422 {\tiny$\pm$.006} & 0.424 {\tiny$\pm$.005} & 0.441 {\tiny$\pm$.004} & 0.432 {\tiny$\pm$.001} \\
 & 336 & \textcolor{red}{\textbf{0.395}} {\tiny$\pm$.007} & \textcolor{red}{\textbf{0.423}} {\tiny$\pm$.005} & \textcolor{blue}{\underline{0.397}} {\tiny$\pm$.019} & \textcolor{blue}{\underline{0.424}} {\tiny$\pm$.006} & 0.419 {\tiny$\pm$.007} & 0.441 {\tiny$\pm$.005} & 0.418 {\tiny$\pm$.009} & 0.440 {\tiny$\pm$.006} & 0.413 {\tiny$\pm$.032} & 0.437 {\tiny$\pm$.023} & 0.436 {\tiny$\pm$.007} & 0.432 {\tiny$\pm$.002} \\
 & 720 & \textcolor{red}{\textbf{0.440}} {\tiny$\pm$.019} & \textcolor{red}{\textbf{0.455}} {\tiny$\pm$.010} & \textcolor{blue}{\underline{0.441}} {\tiny$\pm$.029} & \textcolor{blue}{\underline{0.461}} {\tiny$\pm$.007} & 0.477 {\tiny$\pm$.008} & 0.480 {\tiny$\pm$.005} & 0.477 {\tiny$\pm$.018} & 0.485 {\tiny$\pm$.013} & 0.456 {\tiny$\pm$.008} & 0.471 {\tiny$\pm$.005} & 0.471 {\tiny$\pm$.012} & 0.470 {\tiny$\pm$.007} \\
\midrule
\multirow{4}{*}{\raisebox{3pt}{\rotatebox[origin=l]{90}{ETTh2}}}
 & 96  & \textcolor{blue}{\underline{0.281}} {\tiny$\pm$.006} & \textcolor{blue}{\underline{0.339}} {\tiny$\pm$.004} & \textcolor{red}{\textbf{0.271}} {\tiny$\pm$.003} & \textcolor{red}{\textbf{0.338}} {\tiny$\pm$.002} & 0.287 {\tiny$\pm$.009} & 0.345 {\tiny$\pm$.009} & 0.282 {\tiny$\pm$.009} & 0.345 {\tiny$\pm$.004} & 0.290 {\tiny$\pm$.008} & 0.347 {\tiny$\pm$.003} & 0.285 {\tiny$\pm$.008} & 0.352 {\tiny$\pm$.009} \\
 & 192 & 0.334 {\tiny$\pm$.007} & 0.380 {\tiny$\pm$.005} & \textcolor{red}{\textbf{0.330}} {\tiny$\pm$.030} & \textcolor{blue}{\underline{0.374}} {\tiny$\pm$.019} & \textcolor{blue}{\underline{0.331}} {\tiny$\pm$.009} & 0.383 {\tiny$\pm$.009} & 0.336 {\tiny$\pm$.010} & 0.384 {\tiny$\pm$.010} & 0.333 {\tiny$\pm$.018} & \textcolor{red}{\textbf{0.373}} {\tiny$\pm$.010} & 0.355 {\tiny$\pm$.009} & 0.403 {\tiny$\pm$.007} \\
 & 336 & \textcolor{red}{\textbf{0.314}} {\tiny$\pm$.006} & \textcolor{red}{\textbf{0.374}} {\tiny$\pm$.004} & \textcolor{blue}{\underline{0.322}} {\tiny$\pm$.009} & \textcolor{blue}{\underline{0.380}} {\tiny$\pm$.010} & 0.334 {\tiny$\pm$.011} & 0.394 {\tiny$\pm$.010} & 0.323 {\tiny$\pm$.003} & 0.382 {\tiny$\pm$.003} & 0.323 {\tiny$\pm$.008} & 0.383 {\tiny$\pm$.008} & 0.344 {\tiny$\pm$.008} & 0.402 {\tiny$\pm$.007} \\
 & 720 & \textcolor{red}{\textbf{0.362}} {\tiny$\pm$.008} & \textcolor{red}{\textbf{0.415}} {\tiny$\pm$.006} & \textcolor{blue}{\underline{0.367}} {\tiny$\pm$.009} & \textcolor{blue}{\underline{0.418}} {\tiny$\pm$.008} & 0.398 {\tiny$\pm$.029} & 0.444 {\tiny$\pm$.023} & 0.385 {\tiny$\pm$.016} & 0.437 {\tiny$\pm$.015} & 0.384 {\tiny$\pm$.010} & 0.424 {\tiny$\pm$.009} & 0.429 {\tiny$\pm$.006} & 0.459 {\tiny$\pm$.005} \\
\midrule
\multirow{4}{*}{\raisebox{3pt}{\rotatebox[origin=l]{90}{ETTm1}}}
 & 96  & \textcolor{blue}{\underline{0.292}} {\tiny$\pm$.004} & \textcolor{blue}{\underline{0.346}} {\tiny$\pm$.003} & \textcolor{red}{\textbf{0.291}} {\tiny$\pm$.007} & \textcolor{red}{\textbf{0.345}} {\tiny$\pm$.007} & 0.301 {\tiny$\pm$.006} & 0.354 {\tiny$\pm$.007} & 0.300 {\tiny$\pm$.006} & 0.352 {\tiny$\pm$.005} & 0.296 {\tiny$\pm$.008} & 0.347 {\tiny$\pm$.004} & 0.308 {\tiny$\pm$.006} & 0.353 {\tiny$\pm$.002} \\
 & 192 & \textcolor{blue}{\underline{0.330}} {\tiny$\pm$.004} & 0.371 {\tiny$\pm$.002} & \textcolor{red}{\textbf{0.329}} {\tiny$\pm$.004} & \textcolor{blue}{\underline{0.370}} {\tiny$\pm$.003} & 0.331 {\tiny$\pm$.004} & 0.370 {\tiny$\pm$.002} & 0.335 {\tiny$\pm$.006} & 0.375 {\tiny$\pm$.003} & 0.333 {\tiny$\pm$.004} & \textcolor{red}{\textbf{0.370}} {\tiny$\pm$.002} & 0.339 {\tiny$\pm$.003} & 0.375 {\tiny$\pm$.003} \\
 & 336 & \textcolor{red}{\textbf{0.357}} {\tiny$\pm$.002} & \textcolor{red}{\textbf{0.389}} {\tiny$\pm$.001} & \textcolor{blue}{\underline{0.357}} {\tiny$\pm$.003} & \textcolor{blue}{\underline{0.389}} {\tiny$\pm$.002} & 0.367 {\tiny$\pm$.012} & 0.394 {\tiny$\pm$.005} & 0.363 {\tiny$\pm$.005} & 0.394 {\tiny$\pm$.004} & 0.361 {\tiny$\pm$.002} & 0.390 {\tiny$\pm$.001} & 0.371 {\tiny$\pm$.002} & 0.395 {\tiny$\pm$.003} \\
 & 720 & \textcolor{red}{\textbf{0.409}} {\tiny$\pm$.003} & \textcolor{red}{\textbf{0.422}} {\tiny$\pm$.002} & \textcolor{blue}{\underline{0.410}} {\tiny$\pm$.006} & \textcolor{blue}{\underline{0.423}} {\tiny$\pm$.005} & 0.418 {\tiny$\pm$.009} & 0.426 {\tiny$\pm$.006} & 0.416 {\tiny$\pm$.003} & 0.427 {\tiny$\pm$.006} & 0.414 {\tiny$\pm$.002} & 0.425 {\tiny$\pm$.003} & 0.422 {\tiny$\pm$.005} & 0.425 {\tiny$\pm$.002} \\
\midrule
\multirow{4}{*}{\raisebox{3pt}{\rotatebox[origin=l]{90}{ETTm2}}}
 & 96  & \textcolor{red}{\textbf{0.168}} {\tiny$\pm$.002} & \textcolor{blue}{\underline{0.257}} {\tiny$\pm$.003} & \textcolor{blue}{\underline{0.168}} {\tiny$\pm$.003} & 0.257 {\tiny$\pm$.002} & 0.170 {\tiny$\pm$.003} & \textcolor{red}{\textbf{0.256}} {\tiny$\pm$.001} & 0.172 {\tiny$\pm$.002} & 0.258 {\tiny$\pm$.002} & 0.169 {\tiny$\pm$.003} & 0.257 {\tiny$\pm$.004} & 0.175 {\tiny$\pm$.002} & 0.261 {\tiny$\pm$.001} \\
 & 192 & \textcolor{red}{\textbf{0.225}} {\tiny$\pm$.004} & \textcolor{red}{\textbf{0.297}} {\tiny$\pm$.003} & \textcolor{blue}{\underline{0.228}} {\tiny$\pm$.002} & \textcolor{blue}{\underline{0.299}} {\tiny$\pm$.001} & 0.232 {\tiny$\pm$.006} & 0.299 {\tiny$\pm$.002} & 0.231 {\tiny$\pm$.005} & 0.299 {\tiny$\pm$.004} & 0.229 {\tiny$\pm$.004} & 0.300 {\tiny$\pm$.002} & 0.243 {\tiny$\pm$.002} & 0.310 {\tiny$\pm$.002} \\
 & 336 & \textcolor{red}{\textbf{0.281}} {\tiny$\pm$.007} & \textcolor{blue}{\underline{0.335}} {\tiny$\pm$.006} & \textcolor{blue}{\underline{0.283}} {\tiny$\pm$.012} & 0.335 {\tiny$\pm$.007} & 0.287 {\tiny$\pm$.007} & \textcolor{red}{\textbf{0.334}} {\tiny$\pm$.005} & 0.289 {\tiny$\pm$.006} & 0.338 {\tiny$\pm$.005} & 0.297 {\tiny$\pm$.015} & 0.343 {\tiny$\pm$.009} & 0.299 {\tiny$\pm$.006} & 0.345 {\tiny$\pm$.004} \\
 & 720 & \textcolor{blue}{\underline{0.375}} {\tiny$\pm$.007} & 0.394 {\tiny$\pm$.005} & \textcolor{red}{\textbf{0.374}} {\tiny$\pm$.009} & \textcolor{red}{\textbf{0.391}} {\tiny$\pm$.005} & 0.379 {\tiny$\pm$.002} & 0.394 {\tiny$\pm$.004} & 0.382 {\tiny$\pm$.011} & \textcolor{blue}{\underline{0.394}} {\tiny$\pm$.006} & 0.381 {\tiny$\pm$.007} & 0.405 {\tiny$\pm$.006} & 0.418 {\tiny$\pm$.008} & 0.420 {\tiny$\pm$.005} \\
\midrule
\multirow{4}{*}{\raisebox{3pt}{\rotatebox[origin=l]{90}{Weather}}}
 & 96  & \textcolor{red}{\textbf{0.149}} {\tiny$\pm$.002} & \textcolor{red}{\textbf{0.203}} {\tiny$\pm$.003} & \textcolor{blue}{\underline{0.153}} {\tiny$\pm$.004} & \textcolor{blue}{\underline{0.206}} {\tiny$\pm$.004} & 0.155 {\tiny$\pm$.005} & 0.208 {\tiny$\pm$.006} & 0.157 {\tiny$\pm$.003} & 0.210 {\tiny$\pm$.002} & 0.155 {\tiny$\pm$.003} & 0.207 {\tiny$\pm$.003} & 0.155 {\tiny$\pm$.001} & 0.208 {\tiny$\pm$.001} \\
 & 192 & \textcolor{red}{\textbf{0.191}} {\tiny$\pm$.001} & \textcolor{red}{\textbf{0.242}} {\tiny$\pm$.002} & 0.197 {\tiny$\pm$.002} & 0.246 {\tiny$\pm$.001} & 0.199 {\tiny$\pm$.005} & 0.248 {\tiny$\pm$.005} & 0.201 {\tiny$\pm$.004} & 0.246 {\tiny$\pm$.003} & \textcolor{blue}{\underline{0.195}} {\tiny$\pm$.002} & \textcolor{blue}{\underline{0.246}} {\tiny$\pm$.003} & 0.198 {\tiny$\pm$.001} & 0.247 {\tiny$\pm$.002} \\
 & 336 & 0.245 {\tiny$\pm$.002} & \textcolor{red}{\textbf{0.282}} {\tiny$\pm$.003} & \textcolor{blue}{\underline{0.244}} {\tiny$\pm$.002} & \textcolor{blue}{\underline{0.282}} {\tiny$\pm$.003} & 0.245 {\tiny$\pm$.003} & 0.283 {\tiny$\pm$.002} & \textcolor{red}{\textbf{0.244}} {\tiny$\pm$.001} & 0.283 {\tiny$\pm$.003} & 0.246 {\tiny$\pm$.002} & 0.283 {\tiny$\pm$.004} & 0.245 {\tiny$\pm$.002} & 0.284 {\tiny$\pm$.002} \\
 & 720 & \textcolor{red}{\textbf{0.319}} {\tiny$\pm$.003} & \textcolor{red}{\textbf{0.334}} {\tiny$\pm$.003} & 0.321 {\tiny$\pm$.002} & 0.336 {\tiny$\pm$.001} & \textcolor{blue}{\underline{0.320}} {\tiny$\pm$.003} & \textcolor{blue}{\underline{0.335}} {\tiny$\pm$.004} & 0.322 {\tiny$\pm$.004} & 0.336 {\tiny$\pm$.003} & 0.322 {\tiny$\pm$.005} & 0.337 {\tiny$\pm$.003} & 0.322 {\tiny$\pm$.004} & 0.337 {\tiny$\pm$.002} \\
\midrule
\multirow{4}{*}{\raisebox{3pt}{\rotatebox[origin=l]{90}{Elec.}}}
 & 96  & \textcolor{red}{\textbf{0.137}} {\tiny$\pm$.003} & \textcolor{blue}{\underline{0.238}} {\tiny$\pm$.005} & 0.140 {\tiny$\pm$.002} & 0.242 {\tiny$\pm$.003} & 0.138 {\tiny$\pm$.002} & 0.239 {\tiny$\pm$.003} & 0.139 {\tiny$\pm$.001} & 0.238 {\tiny$\pm$.002} & 0.139 {\tiny$\pm$.001} & 0.240 {\tiny$\pm$.003} & \textcolor{blue}{\underline{0.138}} {\tiny$\pm$.001} & \textcolor{red}{\textbf{0.238}} {\tiny$\pm$.002} \\
 & 192 & \textcolor{red}{\textbf{0.153}} {\tiny$\pm$.001} & \textcolor{red}{\textbf{0.252}} {\tiny$\pm$.003} & 0.157 {\tiny$\pm$.003} & 0.258 {\tiny$\pm$.004} & 0.159 {\tiny$\pm$.004} & 0.260 {\tiny$\pm$.003} & 0.159 {\tiny$\pm$.004} & 0.260 {\tiny$\pm$.003} & \textcolor{blue}{\underline{0.155}} {\tiny$\pm$.002} & \textcolor{blue}{\underline{0.255}} {\tiny$\pm$.004} & 0.157 {\tiny$\pm$.000} & 0.257 {\tiny$\pm$.001} \\
 & 336 & \textcolor{blue}{\underline{0.172}} {\tiny$\pm$.002} & \textcolor{blue}{\underline{0.273}} {\tiny$\pm$.003} & 0.173 {\tiny$\pm$.001} & 0.282 {\tiny$\pm$.001} & 0.174 {\tiny$\pm$.003} & 0.275 {\tiny$\pm$.004} & 0.176 {\tiny$\pm$.003} & 0.275 {\tiny$\pm$.002} & \textcolor{red}{\textbf{0.172}} {\tiny$\pm$.002} & \textcolor{red}{\textbf{0.271}} {\tiny$\pm$.004} & 0.176 {\tiny$\pm$.000} & 0.275 {\tiny$\pm$.002} \\
 & 720 & \textcolor{red}{\textbf{0.209}} {\tiny$\pm$.005} & \textcolor{red}{\textbf{0.304}} {\tiny$\pm$.005} & 0.217 {\tiny$\pm$.003} & 0.312 {\tiny$\pm$.004} & 0.223 {\tiny$\pm$.005} & 0.322 {\tiny$\pm$.005} & 0.227 {\tiny$\pm$.002} & 0.324 {\tiny$\pm$.002} & 0.221 {\tiny$\pm$.003} & 0.316 {\tiny$\pm$.002} & \textcolor{blue}{\underline{0.214}} {\tiny$\pm$.002} & \textcolor{blue}{\underline{0.306}} {\tiny$\pm$.001} \\
\midrule
\multirow{4}{*}{\raisebox{3pt}{\rotatebox[origin=l]{90}{Solar}}}
 & 96  & \textcolor{red}{\textbf{0.177}} {\tiny$\pm$.003} & \textcolor{red}{\textbf{0.241}} {\tiny$\pm$.007} & 0.185 {\tiny$\pm$.003} & 0.250 {\tiny$\pm$.007} & 0.187 {\tiny$\pm$.005} & 0.257 {\tiny$\pm$.009} & 0.187 {\tiny$\pm$.002} & 0.261 {\tiny$\pm$.006} & 0.185 {\tiny$\pm$.006} & 0.252 {\tiny$\pm$.012} & \textcolor{blue}{\underline{0.184}} {\tiny$\pm$.006} & \textcolor{blue}{\underline{0.242}} {\tiny$\pm$.009} \\
 & 192 & \textcolor{red}{\textbf{0.199}} {\tiny$\pm$.002} & \textcolor{red}{\textbf{0.265}} {\tiny$\pm$.002} & 0.202 {\tiny$\pm$.003} & 0.271 {\tiny$\pm$.005} & 0.213 {\tiny$\pm$.005} & 0.275 {\tiny$\pm$.008} & 0.207 {\tiny$\pm$.003} & 0.270 {\tiny$\pm$.002} & \textcolor{blue}{\underline{0.202}} {\tiny$\pm$.003} & 0.267 {\tiny$\pm$.009} & 0.208 {\tiny$\pm$.009} & \textcolor{blue}{\underline{0.266}} {\tiny$\pm$.005} \\
 & 336 & \textcolor{red}{\textbf{0.210}} {\tiny$\pm$.001} & \textcolor{blue}{\underline{0.272}} {\tiny$\pm$.001} & \textcolor{blue}{\underline{0.210}} {\tiny$\pm$.005} & 0.276 {\tiny$\pm$.002} & 0.220 {\tiny$\pm$.006} & 0.285 {\tiny$\pm$.005} & 0.217 {\tiny$\pm$.007} & 0.278 {\tiny$\pm$.008} & 0.216 {\tiny$\pm$.008} & 0.276 {\tiny$\pm$.006} & 0.214 {\tiny$\pm$.003} & \textcolor{red}{\textbf{0.269}} {\tiny$\pm$.002} \\
 & 720 & \textcolor{red}{\textbf{0.210}} {\tiny$\pm$.000} & 0.276 {\tiny$\pm$.002} & \textcolor{blue}{\underline{0.210}} {\tiny$\pm$.002} & \textcolor{red}{\textbf{0.273}} {\tiny$\pm$.004} & 0.228 {\tiny$\pm$.009} & 0.293 {\tiny$\pm$.009} & 0.226 {\tiny$\pm$.007} & 0.302 {\tiny$\pm$.002} & 0.224 {\tiny$\pm$.005} & 0.292 {\tiny$\pm$.005} & 0.217 {\tiny$\pm$.006} & \textcolor{blue}{\underline{0.276}} {\tiny$\pm$.006} \\
\midrule
\multirow{4}{*}{\raisebox{3pt}{\rotatebox[origin=l]{90}{Traffic}}}
 & 96  & \textcolor{red}{\textbf{0.384}} {\tiny$\pm$.002} & \textcolor{red}{\textbf{0.269}} {\tiny$\pm$.003} & \textcolor{blue}{\underline{0.386}} {\tiny$\pm$.002} & 0.276 {\tiny$\pm$.002} & 0.388 {\tiny$\pm$.013} & 0.278 {\tiny$\pm$.011} & 0.394 {\tiny$\pm$.007} & 0.283 {\tiny$\pm$.005} & 0.389 {\tiny$\pm$.007} & 0.277 {\tiny$\pm$.006} & 0.403 {\tiny$\pm$.002} & \textcolor{blue}{\underline{0.272}} {\tiny$\pm$.003} \\
 & 192 & \textcolor{red}{\textbf{0.403}} {\tiny$\pm$.003} & \textcolor{red}{\textbf{0.281}} {\tiny$\pm$.001} & \textcolor{blue}{\underline{0.406}} {\tiny$\pm$.005} & 0.283 {\tiny$\pm$.004} & 0.422 {\tiny$\pm$.009} & 0.291 {\tiny$\pm$.006} & 0.406 {\tiny$\pm$.006} & \textcolor{blue}{\underline{0.281}} {\tiny$\pm$.010} & 0.415 {\tiny$\pm$.003} & 0.291 {\tiny$\pm$.003} & 0.435 {\tiny$\pm$.001} & 0.306 {\tiny$\pm$.007} \\
 & 336 & \textcolor{red}{\textbf{0.419}} {\tiny$\pm$.008} & \textcolor{red}{\textbf{0.286}} {\tiny$\pm$.008} & \textcolor{blue}{\underline{0.420}} {\tiny$\pm$.002} & \textcolor{blue}{\underline{0.289}} {\tiny$\pm$.003} & 0.435 {\tiny$\pm$.006} & 0.305 {\tiny$\pm$.006} & 0.438 {\tiny$\pm$.010} & 0.299 {\tiny$\pm$.006} & 0.429 {\tiny$\pm$.005} & 0.303 {\tiny$\pm$.007} & 0.448 {\tiny$\pm$.002} & 0.316 {\tiny$\pm$.002} \\
 & 720 & \textcolor{blue}{\underline{0.459}} {\tiny$\pm$.003} & \textcolor{blue}{\underline{0.309}} {\tiny$\pm$.003} & \textcolor{red}{\textbf{0.449}} {\tiny$\pm$.002} & \textcolor{red}{\textbf{0.305}} {\tiny$\pm$.002} & 0.488 {\tiny$\pm$.010} & 0.331 {\tiny$\pm$.008} & 0.517 {\tiny$\pm$.005} & 0.334 {\tiny$\pm$.007} & 0.473 {\tiny$\pm$.009} & 0.325 {\tiny$\pm$.002} & 0.559 {\tiny$\pm$.012} & 0.315 {\tiny$\pm$.005} \\
\midrule
\multicolumn{2}{c|}{\textbf{1st-place count}}
 & \textbf{24} & \textbf{20} & 6 & 5 & 0 & 2 & 1 & 0 & 1 & 3 & 0 & 2 \\
\bottomrule
\end{tabular}%
}
\end{table*}

\begin{figure}[ht!]
    \centering
    \includegraphics[width=0.98\linewidth]{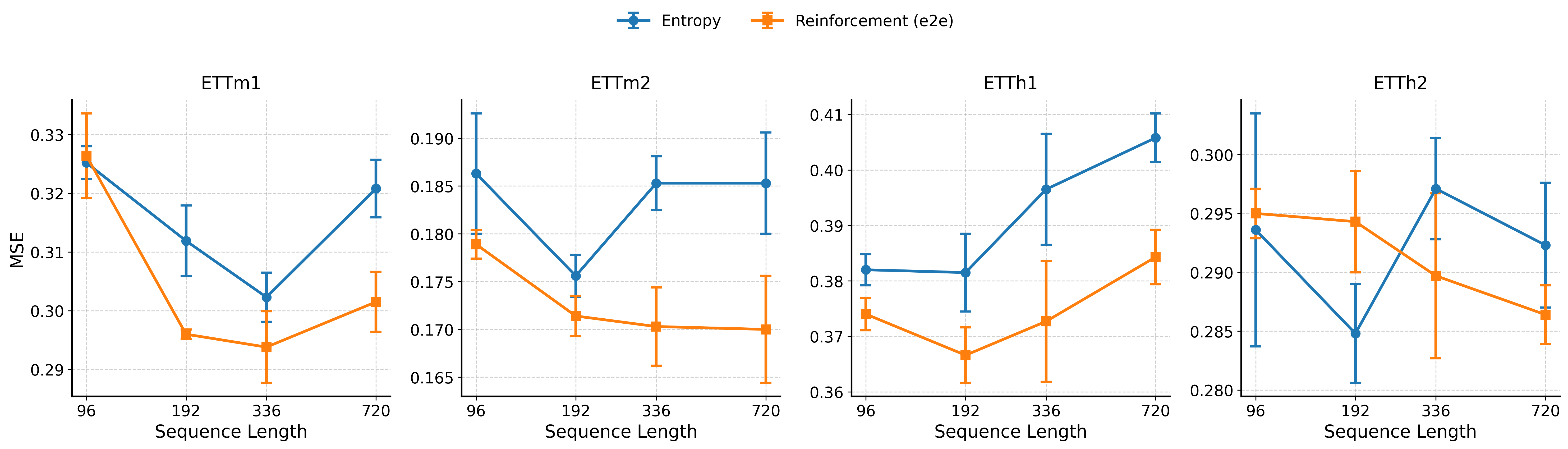}
    \caption{MSE at forecasting horizon $H=96$ across look-back windows. Entropy patching is competitive for short contexts but struggles as the input context expands, whereas \model maintains strong performance across most datasets and context lengths.}
    \label{fig:ablation_seqlen}
\end{figure}

\textbf{Hyperparameter ablation.} It is a well-known phenomenon in time series forecasting that performance can often degrade as the look-back window continues to increase, as models struggle to extract robust trends from the accumulated noise~\citep{zeng2023transformers}. Therefore, adaptive patching must remain effective as context grows. We compare \model with entropy patching~\cite{pagnoni2025byte,abeywickrama2025entrope} across $T \in \{96, 192, 336, 720\}$ on $H=96$. As Figure~\ref{fig:ablation_seqlen} shows, both methods are competitive on short sequence length $T$, where entropy provides competitive performance relative to \model. However, \model consistently pulls ahead on large $T$, scaling better as context window increases.

\begin{figure}
    \centering
    \includegraphics[width=0.98\linewidth]{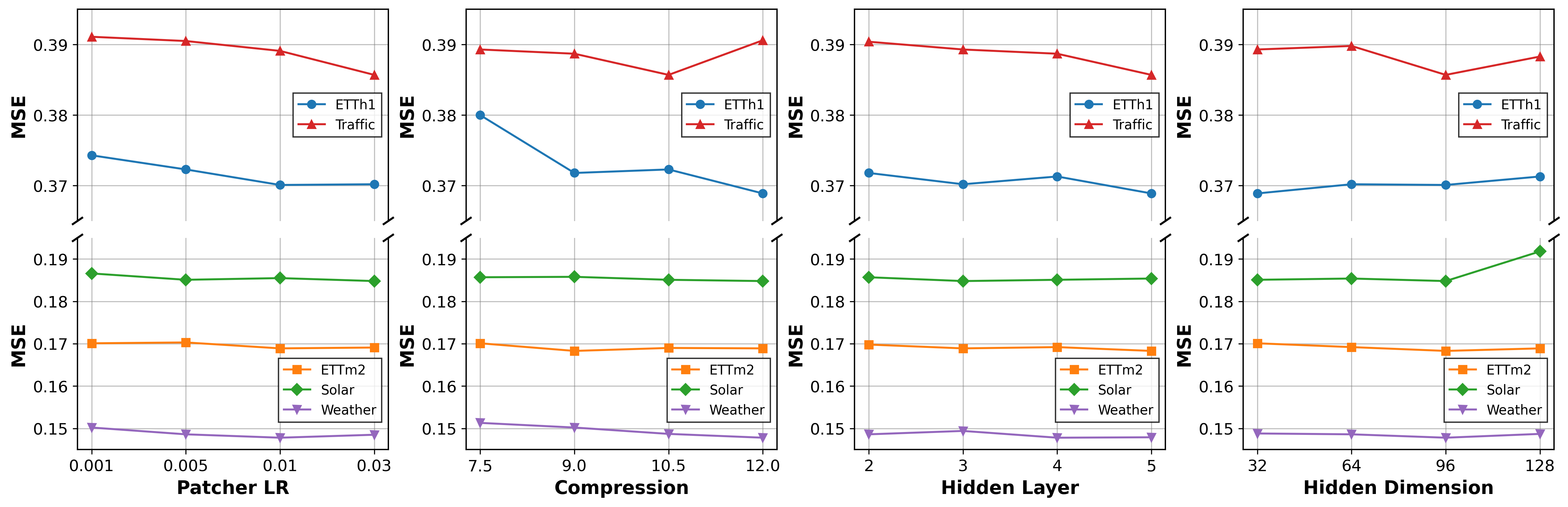}
    \caption{Hyperparameter sensitivity analysis at forecasting horizon $H=96$; lower MSE is better. From left to right, we vary the patcher learning rate, the minimum compression rate $r_m$, and the backbone's number of hidden layers and hidden dimension.}
    \label{fig:params_sensitivity} 
\end{figure}

This scaling trend highlights a limitation of boundary policies driven by auxiliary metrics such as entropy: the proxy can work for short, simple contexts, but longer windows introduce more heterogeneous patterns and widen the mismatch between estimated entropy and forecasting utility. In contrast, \model learns boundaries end-to-end for the forecasting objective. The widening gap therefore supports trainable patching as a way to compress long contexts while preserving the temporal detail most useful for prediction. 
% We also note that H-Net, the prior end-to-end patching baseline using the same backbone architecture as in \citet{hwang2025dynamic}, remains weak in MSE: it has no first-place MSE entries in Table~\ref{tab:results_full} and gives the worst average MSE on ETTh1 and Traffic in Table~\ref{tab:results_avg}.
Other than look-back window, our method is not particularly sensitive to hyperparameters, see Figure~\ref{fig:params_sensitivity} for details.

\textbf{Efficiency analysis.} Table~\ref{tab:inference_efficiency_336_720_32} reports inference-only memory and throughput on ETTh1 with $L=336$, $H=720$, and batch size $32$. As expected, static patching remains the cheapest baseline because it does not perform input-dependent boundary selection. Nevertheless, \model incurs only a moderate overhead over static patching, increasing memory from 111.9 to 138.6 MB/sample and reducing throughput from 949.8 to 847.0 samples/s, while retaining the adaptive segmentation behavior that improves forecasting accuracy. Compared with selective patching, \model achieves higher throughput despite using slightly more memory. More importantly, \model is substantially more memory-efficient than entropy patching, reducing memory by more than half (138.6 vs.\ 323.4 MB/sample) while maintaining comparable throughput. These results indicate that the learned patching policy provides a practical efficiency--accuracy trade-off: it preserves the benefits of adaptive tokenization without introducing the large inference footprint of auxiliary-model-based patching. A full breakdown of trainable, total, and patcher parameters together with MACs per method is provided in Table~\ref{tab:compute_complexity} of Appendix~\ref{appendix:compute-complexity}.

% Inference-only efficiency table (seq len 336, pred len 720, batch size 32)
\begin{table*}[ht!]
\centering
\begin{minipage}[t]{0.32\textwidth}
    \centering
    \captionsetup{justification=raggedright,singlelinecheck=false}
    \caption{ETTh1 inference efficiency ($L=336$, $H=720$).}
    \label{tab:inference_efficiency_336_720_32}
    {\scriptsize
    \renewcommand{\arraystretch}{1.40}
    \setlength{\tabcolsep}{0.8pt}
    \begin{tabular}{@{}lcc@{}}
    \toprule
    \textbf{Method} & \shortstack{\textbf{Memory}\\\textbf{(MB/sample)}} & \shortstack{\textbf{Throughput}\\\textbf{(samples/s)}} \\
    \midrule
    Static~\cite{nie2023patch} & 111.9 & 949.8 \\
    Selective~\cite{wuenhancing} & 113.3 & 697.6 \\
    \model\ [Ours] & 138.6 & 847.0 \\
    Entropy~\cite{abeywickrama2025entrope} & 323.4 & 883.6 \\
    % H-Net's Chunking~\cite{hwang2025dynamic} & 490.4 & 3097.7 \\
    \bottomrule
    \end{tabular}}
\end{minipage}
\hfill
\begin{minipage}[t]{0.61\textwidth}
    \centering
    \caption{Pairwise permutation-test $p$-values for fixed-rate boundary-location distributions with target compression rate $r_c=8$.}
    \label{tab:fixed_rate_pairwise_pvalues}
    {\scriptsize
    \renewcommand{\arraystretch}{1.0}
    \setlength{\tabcolsep}{2.5pt}
    \resizebox{\linewidth}{!}{%
    \begin{tabular}{lcccccccc}
        \toprule
         & \textbf{ETTh1} & \textbf{ETTh2} & \textbf{ETTm1} & \textbf{ETTm2} & \textbf{Elec.} & \textbf{Weath.} & \textbf{Solar} & \textbf{Traffic} \\
        \midrule
        \textbf{ETTh1} & -- & $<0.001$ & \textbf{0.019} & $<0.001$ & $<0.001$ & $<0.001$ & $<0.001$ & $<0.001$ \\
        \textbf{ETTh2} &    & --       & \textbf{0.043} & \textbf{0.265} & $<0.001$ & $<0.001$ & $<0.001$ & $<0.001$ \\
        \textbf{ETTm1} &    &          & --     & 0.003   & $<0.001$ & $<0.001$ & $<0.001$ & $<0.001$ \\
        \textbf{ETTm2} &    &          &        & --       & $<0.001$ & $<0.001$ & $<0.001$ & $<0.001$ \\
        \textbf{Elec.} &    &          &        &          & --       & $<0.001$ & $<0.001$ & $<0.001$ \\
        \textbf{Weath.}&    &          &        &          &          & --       & $<0.001$ & $<0.001$ \\
        \textbf{Solar} &    &          &        &          &          &          & --       & $<0.001$ \\
        \textbf{Traffic} &  &          &        &          &          &          &          & -- \\
        \bottomrule
    \end{tabular}}}
\end{minipage}
\end{table*}

\textbf{Statistical evidence of data adaptivity.} The foundation patcher achieves an average compression rate of $10.92$ on UTSD, comfortably above the minimum of $8$, indicating that it has internalized the compression constraint and allocates boundaries selectively through pre-training. However, we want to further quantify whether the learned foundation patcher changes \emph{where} it places boundaries across datasets with different characteristics, to investigate whether the patcher has learned specific patterns or is only setting semi-random boundaries to satisfy compression rate. We run the patcher in fixed-rate inference mode with target compression rate $r_c=8$, which fixes the boundary budget and removes patch count as the main explanation for dataset differences. For each dataset, we estimate a boundary-frequency curve from sampled windows after excluding the forced initial boundary, and compare datasets by the squared distance between their empirical curves. Pairwise significance is assessed with 1000 label permutations within each dataset pair while preserving sample counts; at this resolution, $p=9.99{\times}10^{-4}$ means that no permutation produced a difference at least as large as the observed one.

Table~\ref{tab:fixed_rate_pairwise_pvalues} summarizes the resulting fixed-rate boundary-location tests across eight datasets. Overall, 27 of 28 dataset pairs are significant at $\alpha=0.05$. Most pairs reach the minimum attainable Monte Carlo $p$-value, while the closest significant ETT pair is ETTh2--ETTm1 ($p=0.043$). The only nonsignificant pair is ETTh2--ETTm2 ($p=0.265$), indicating similar boundary-location distributions for these two datasets under the fixed-rate control. Because the fixed-rate setting controls the boundary budget, the broad pattern of significant differences shows that \model adapts boundary locations to dataset structure rather than resorting to trivial patterns that satisfy compression rate.

\begin{figure}[!ht]
    \centering
    \includegraphics[width=0.98\linewidth]{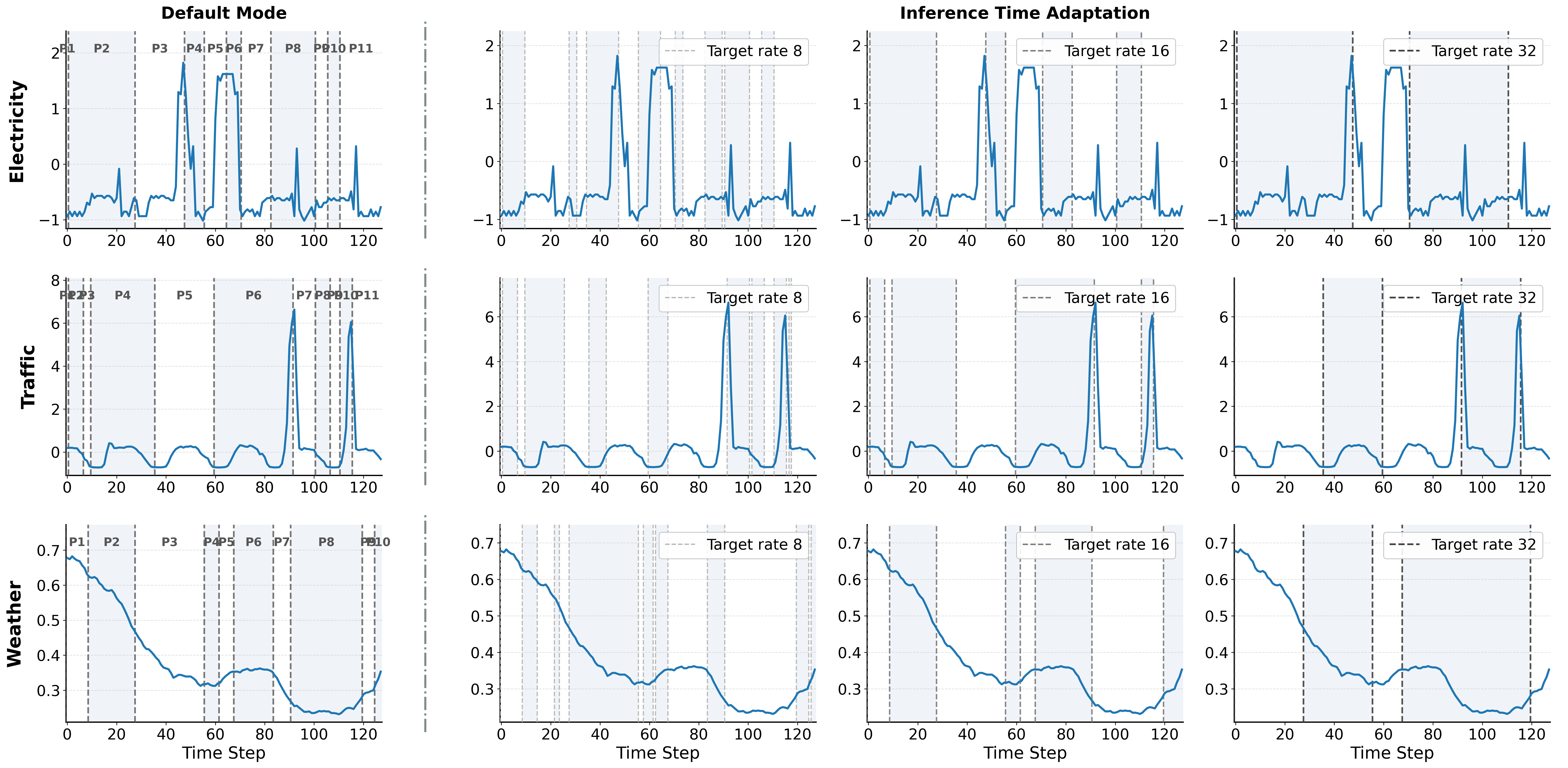}
    \caption{Representative patch segmentations sampled from the foundation patcher trained with a minimum compression rate of 8 on UTSD. The learned boundaries naturally concentrate around major transitions while keeping short neighborhoods of local temporal structure together.}
    \label{fig:patch_samples}
\end{figure}
\textbf{Interpreting the patching policy.}
In Figure~\ref{fig:patch_samples}, we visualize the segmentations learned by \model under the foundation patcher setting on Electricity, Traffic, and Weather. Consistent with some previous heuristics~\cite{pagnoni2025byte, hwang2025dynamic, ankireddytimesqueeze}, boundaries mainly concentrate near major turning points; however, \model also \textit{wraps} minor shifts within local neighborhood that preserve seasonality to some extent (see electricity and traffic for example), likely to preserve informative transition shapes and avoid reducing patches to less informative monotone fragments. Furthermore, the policy appears to account for compression loss over long windows by keeping patches structurally compact. It rarely spans more than a few major turning points, and still places boundaries during a single trend if the same trend persists for an extended period (see the smoother curve on the weather dataset for example).
%The patches remain compact, rarely spanning more than a few major turning points, which balances aggressive compression with the within-patch variation needed by the downstream forecaster. 

These patterns help explain why \model consistently outperforms H-Net, our other end-to-end patching baseline. H-Net was designed for byte-level sequence modeling and places boundaries using cosine similarity between high-dimensional Mamba embeddings at full resolution. For continuous time series, however, useful boundaries should compress the raw trajectory while preserving local transitions and salient events; embedding-space similarity need not align with these raw-signal structures. Thus, H-Net's geometry-driven inductive bias can produce coherent chunks for discrete sequences while missing local structures that matter for forecasting.

\section{Conclusion}
\label{sec:conclusion}

We introduced Reinforcement Patching (\model), a novel framework that treats data-adaptive sequence representation as a discrete optimization problem. By utilizing a contextual bandit formulation and Group Relative Policy Gradient (GRPG), \model bypasses soft-discretization relaxations and cleanly decouples the patching policy from the downstream sequence backbone. Experiments in multivariate time series forecasting demonstrate that \model consistently outperforms state-of-the-art patching strategies in both joint end-to-end training and zero-shot transfer settings. It scales robustly to extended context windows and provides the community with empirical insights into naturally learned, performance-driven segmentation patterns.

\textbf{Limitations.} First, despite its impressive transferability, our foundation patcher was pre-trained on univariate dynamics. Large-scale, multivariate pre-training to capture cross-channel patching dependencies remains an open challenge within the broader time series modeling umbrella. Second, while the unified one-step MDP formulation enables highly efficient parallelized training, letting the patcher further contextualize its decisions made on-the-fly in the multi-step, causal MDP formulation could still be desirable. Efficiently training deep models in this formulation remains an open challenge.

\textbf{Future Work.} Because the fundamental principles of ReinPatch are domain-agnostic, we plan to explore its potential application as a dynamic, tokenization-free alternative to BPE in large language models, as well as in audio and genomics. This includes adapting RL objectives (e.g., PPO) for settings where the downstream backbone is a massive-scale LLM, and exploring task-specific fine-tuning of pre-trained patching policies. Besides, since the RL formulation is reward-agnostic, the forecasting reward can likewise be replaced with a self-supervised objective (e.g., masked-patch reconstruction), turning the patcher into a self-supervised segmentation policy and a dynamic front-end that feeds variable-length patches to an SSL-pretrained backbone.

\vfill
\pagebreak

\begin{ack}
  We thank Doron Bergman, Rizal Fathony, Nihal Sharma, and Thomas Caputo for the valuable discussions.
\end{ack}

\bibliography{references}

@article{hwang2025dynamic,
  title={{Dynamic chunking for end-to-end hierarchical sequence modeling}},
  author={Hwang, Sukjun and Wang, Brandon and Gu, Albert},
  journal={arXiv preprint arXiv:2507.07955},
  year={2025}
}

@article{abeywickrama2025entrope,
  title={{Entropy Guided Dynamic Patch Segmentation for Time Series Transformers}},
  author={Abeywickrama, Sachith and Eldele, Emadeldeen and Wu, Min and Li, Xiaoli and Yuen, Chau},
  journal={arXiv preprint arXiv:2509.26157},
  year={2025}
}

@inproceedings{ankireddytimesqueeze,
  title={{TimeSqueeze: Dynamic Patching for Efficient Time Series Forecasting}},
  author={Ankireddy, Sravan Kumar and Seleznev, Nikita and Nguyen, Nam H and Wu, Yulun and Kumar, Senthil and Huang, Furong and Bruss, C Bayan},
  booktitle={Recent Advances in Time Series Foundation Models Have We Reached the'BERT Moment'?},
  year={2025}
}

@inproceedings{wuenhancing,
  title={{Enhancing Time Series Forecasting through Selective Representation Spaces: A Patch Perspective}},
  author={Wu, Xingjian and Qiu, Xiangfei and Cheng, Hanyin and Li, Zhengyu and Hu, Jilin and Guo, Chenjuan and Yang, Bin},
  booktitle={The Thirty-ninth Annual Conference on Neural Information Processing Systems},
  year={2025}
}

@inproceedings{wang2025lightgts,
  title={{LightGTS: A Lightweight General Time Series Forecasting Model}},
  author={Wang, Yihang and Qiu, Yuying and Chen, Peng and Shu, Yang and Rao, Zhongwen and Pan, Lujia and Yang, Bin and Guo, Chenjuan},
  booktitle={International Conference on Machine Learning},
  pages={64109--64126},
  year={2025},
  organization={PMLR}
}

@inproceedings{nie2023patch,
  title     = {{A Time Series is Worth 64 Words: Long-term Forecasting with Transformers}},
  author    = {Nie, Yuqi and
               H. Nguyen, Nam and
               Sinthong, Phanwadee and 
               Kalagnanam, Jayant},
  booktitle = {International Conference on Learning Representations},
  year      = {2023}
}

@inproceedings{sennrich2016neural,
  title={{Neural machine translation of rare words with subword units}},
  author={Sennrich, Rico and Haddow, Barry and Birch, Alexandra},
  booktitle={Proceedings of the 54th annual meeting of the association for computational linguistics (volume 1: long papers)},
  pages={1715--1725},
  year={2016}
}

@inproceedings{zhou2021informer,
  title={{Informer: Beyond efficient transformer for long sequence time-series forecasting}},
  author={Zhou, Haoyi and Zhang, Shanghang and Peng, Jieqi and Zhang, Shuai and Li, Jianxin and Xiong, Hui and Zhang, Wancai},
  booktitle={Proceedings of the AAAI conference on artificial intelligence},
  volume={35},
  pages={11106--11115},
  year={2021}
}

@article{wu2021autoformer,
  title={{Autoformer: Decomposition transformers with auto-correlation for long-term series forecasting}},
  author={Wu, Haixu and Xu, Jiehui and Wang, Jianmin and Long, Mingsheng},
  journal={Advances in neural information processing systems},
  volume={34},
  pages={22419--22430},
  year={2021}
}

@inproceedings{zhou2022fedformer,
  title={{Fedformer: Frequency enhanced decomposed transformer for long-term series forecasting}},
  author={Zhou, Tian and Ma, Ziqing and Wen, Qingsong and Wang, Xue and Sun, Liang and Jin, Rong},
  booktitle={International Conference on Machine Learning},
  pages={27268--27286},
  year={2022},
  organization={PMLR}
}

@inproceedings{dosovitskiy2021image,
  title={{An Image is Worth 16x16 Words: Transformers for Image Recognition at Scale}},
  author={Dosovitskiy, Alexey and Beyer, Lucas and Kolesnikov, Alexander and Weissenborn, Dirk and Zhai, Xiaohua and Unterthiner, Thomas and Dehghani, Mostafa and Minderer, Matthias and Heigold, Georg and Gelly, Sylvain and others},
  booktitle={International Conference on Learning Representations},
  year={2021},
  organization={PMLR}
}

@inproceedings{das2024decoder,
  title={{A decoder-only foundation model for time-series forecasting}},
  author={Das, Abhimanyu and Kong, Weihao and Sen, Rajat and Zhou, Yichen},
  booktitle={Forty-first International Conference on Machine Learning},
  year={2024}
}

@article{goswami2024moment,
  title={{Moment: A family of open time-series foundation models}},
  author={Goswami, Mononito and Szafer, Konrad and Choudhry, Arjun and Cai, Yifu and Li, Shuo and Dubrawski, Artur},
  journal={arXiv preprint arXiv:2402.03885},
  year={2024}
}

@inproceedings{woo2024unified,
  title={{Unified training of universal time series forecasting transformers}},
  author={Woo, Gerald and Liu, Chenghao and Kumar, Akshat and Xiong, Caiming and Savarese, Silvio and Sahoo, Doyen},
  booktitle={Forty-first International Conference on Machine Learning},
  year={2024}
}

@inproceedings{liu2024timer,
  title={{Timer-XL: Long-Context Transformers for Unified Time Series Forecasting}},
  author={Liu, Yong and Qin, Guo and Huang, Xiangdong and Wang, Jianmin and Long, Mingsheng},
  booktitle={The Thirteenth International Conference on Learning Representations},
  year={2025}
}

@inproceedings{huang2024hdmixer,
  title={{Hdmixer: Hierarchical dependency with extendable patch for multivariate time series forecasting}},
  author={Huang, Qihe and Shen, Lei and Zhang, Ruixin and Cheng, Jiahuan and Ding, Shouhong and Zhou, Zhengyang and Wang, Yang},
  booktitle={Proceedings of the AAAI conference on artificial intelligence},
  volume={38},
  pages={12608--12616},
  year={2024}
}

@inproceedings{zhang2024irregular,
  title={{Irregular multivariate time series forecasting: A transformable patching graph neural networks approach}},
  author={Zhang, Weijia and Yin, Chenlong and Liu, Hao and Zhou, Xiaofang and Xiong, Hui},
  booktitle={Forty-first International Conference on Machine Learning},
  year={2024}
}

@inproceedings{ronneberger2015u,
  title={{U-net: Convolutional networks for biomedical image segmentation}},
  author={Ronneberger, Olaf and Fischer, Philipp and Brox, Thomas},
  booktitle={International Conference on Medical image computing and computer-assisted intervention},
  pages={234--241},
  year={2015},
  organization={Springer}
}

@inproceedings{pagnoni2025byte,
  title={{Byte latent transformer: Patches scale better than tokens}},
  author={Pagnoni, Artidoro and Pasunuru, Ramakanth and Rodriguez, Pedro and Nguyen, John and Muller, Benjamin and Li, Margaret and Zhou, Chunting and Yu, Lili and Weston, Jason E and Zettlemoyer, Luke and others},
  booktitle={Proceedings of the 63rd Annual Meeting of the Association for Computational Linguistics (Volume 1: Long Papers)},
  pages={9238--9258},
  year={2025}
}

@article{slagle2024spacebyte,
  title={{Spacebyte: Towards deleting tokenization from large language modeling}},
  author={Slagle, Kevin},
  journal={Advances in Neural Information Processing Systems},
  volume={37},
  pages={124925--124950},
  year={2024}
}

@inproceedings{nawrot2023efficient,
  title={{Efficient transformers with dynamic token pooling}},
  author={Nawrot, Piotr and Chorowski, Jan and Lancucki, Adrian and Ponti, Edoardo Maria},
  booktitle={Proceedings of the 61st Annual Meeting of the Association for Computational Linguistics (Volume 1: Long Papers)},
  pages={6403--6417},
  year={2023}
}

@article{lecun2015deep,
  title={{Deep learning}},
  author={LeCun, Yann and Bengio, Yoshua and Hinton, Geoffrey},
  journal={nature},
  volume={521},
  number={7553},
  pages={436--444},
  year={2015},
  publisher={Nature Publishing Group UK London}
}

@article{shao2024deepseekmath,
  title={{Deepseekmath: Pushing the limits of mathematical reasoning in open language models}},
  author={Shao, Zhihong and Wang, Peiyi and Zhu, Qihao and Xu, Runxin and Song, Junxiao and Bi, Xiao and Zhang, Haowei and Zhang, Mingchuan and Li, YK and others},
  journal={arXiv preprint arXiv:2402.03300},
  year={2024}
}

@inproceedings{liutimer,
  title={{Timer: Generative Pre-trained Transformers Are Large Time Series Models}},
  author={Liu, Yong and Zhang, Haoran and Li, Chenyu and Huang, Xiangdong and Wang, Jianmin and Long, Mingsheng},
  booktitle={Forty-first International Conference on Machine Learning},
  year={2024}
}

@inproceedings{zeng2023transformers,
  title={{Are transformers effective for time series forecasting?}},
  author={Zeng, Ailing and Chen, Muxi and Zhang, Lei and Xu, Qiang},
  booktitle={Proceedings of the AAAI conference on artificial intelligence},
  volume={37},
  pages={11121--11128},
  year={2023}
}

@article{schulman2017proximal,
  title={{Proximal policy optimization algorithms}},
  author={Schulman, John and Wolski, Filip and Dhariwal, Prafulla and Radford, Alec and Klimov, Oleg},
  journal={arXiv preprint arXiv:1707.06347},
  year={2017}
}

@inproceedings{mnih2016asynchronous,
  title={{Asynchronous methods for deep reinforcement learning}},
  author={Mnih, Volodymyr and Badia, Adria Puigdomenech and Mirza, Mehdi and Graves, Alex and Lillicrap, Timothy and Harley, Tim and Silver, David and Kavukcuoglu, Koray},
  booktitle={International conference on machine learning},
  pages={1928--1937},
  year={2016},
  organization={PmLR}
}

@article{dauncey2026you,
  title={You Can Learn Tokenization End-to-End with Reinforcement Learning},
  author={Dauncey, Sam and Wattenhofer, Roger},
  journal={arXiv preprint arXiv:2602.13940},
  year={2026}
}

@article{paszke2019pytorch,
  title={Pytorch: An imperative style, high-performance deep learning library},
  author={Paszke, Adam and Gross, Sam and Massa, Francisco and Lerer, Adam and Bradbury, James and Chanan, Gregory and Killeen, Trevor and Lin, Zeming and Gimelshein, Natalia and Antiga, Luca and others},
  journal={Advances in neural information processing systems},
  volume={32},
  year={2019}
}

@inproceedings{katariya2026dp,
  title={DP-Rec: Towards Dynamic Patching for Efficient Long-Sequence Recommendation},
  author={Katariya, Dwipam and Caputo, Thomas and Shreemali, Akshat and Origgi, Juan Manuel and Seleznev, Nikita and Mohanty, Pranab and Mishra, Kalanand and Nguyen, Nam and Montgomery, James},
  booktitle={Proceedings of the 20th ACM Conference on Recommender Systems},
  pages={182--191},
  year={2026}
}

@article{friedman1937use,
  title={The use of ranks to avoid the assumption of normality implicit in the analysis of variance},
  author={Friedman, Milton},
  journal={Journal of the american statistical association},
  volume={32},
  number={200},
  pages={675--701},
  year={1937},
  publisher={Taylor \& Francis}
}

@book{nemenyi1963distribution,
  title={Distribution-free multiple comparisons.},
  author={Nemenyi, Peter Bjorn},
  year={1963},
  publisher={Princeton University}
}

@article{demvsar2006statistical,
  title={Statistical comparisons of classifiers over multiple data sets},
  author={Dem{\v{s}}ar, Janez},
  journal={Journal of Machine learning research},
  volume={7},
  number={Jan},
  pages={1--30},
  year={2006}
}
\bibliographystyle{references}

\newpage

\appendix
\section*{Appendix}

\section{Extended Related Work}
\label{appendix:extended_related_work}

\textbf{Static patching.}
Introduced in PatchTST~\citep{nie2023patch}, patch-based compression has become a central technique for scaling transformer models for time series. %By embedding contiguous sub-sequences, or \textit{patches}, rather than individual time steps, patching reduces the effective sequence length while preserving local temporal structure. 
This design has since been adopted by several time-series foundation models, including TimesFM~\citep{das2024decoder}, MOMENT~\citep{goswami2024moment}, Moirai~\citep{woo2024unified}, and Timer-XL~\citep{liu2024timer}, collectively demonstrating that patching substantially improves training and inference efficiency. However, these approaches rely on a static patch size for an entire input sequence, which limits their ability to model real-world series with substantial temporal heterogeneity%, highlighting the need for adaptive compression strategies that can vary patch granularity according to the local structure of the signal
.

\textbf{Data-adaptive patching.}
Several recent methods have explored adapting patch size automatically rather than selecting it through hand-crafted heuristics. HDMixer~\citep{huang2024hdmixer} enlarges the receptive field by selectively combining adjacent fixed-size patches. IMTS~\citep{zhang2024irregular} is designed for irregularly sampled series and adjusts patch size to maintain a consistent number of observations per patch. LightGTS~\citep{wang2025lightgts} selects patch sizes for each input based on periodicity in the Fourier domain, while SRSNet~\citep{wuenhancing} dynamically selects and combines features from several fixed-size patches to improve forecasting performance. Although these approaches introduce adaptivity, they do not explicitly learn within-sequence patching policies that vary patch sizes inside a single series according to local signal characteristics. Furthermore, SRSNet still includes a static patching component while using adaptive patches as auxiliary signal.

\textbf{Patching using external rules or models.}
In language modeling, there has been growing interest in moving beyond static offline tokenization schemes such as BPE. Since conventional tokenization can introduce systematic biases and brittle dependencies, several recent models instead operate directly on bytes. %However, naïve byte-level processing leads to prohibitively long input sequences~\citep{slagle2024spacebyte}, making efficient compression essential. To address this challenge, 
Prior work has proposed adaptive segmentation methods that rely on external identifiers or auxiliary models to determine patch boundaries: SpaceByte~\citep{slagle2024spacebyte} uses delimiter-like markers, such as spaces, to group bytes into embeddings, while Byte Latent Transformer (BLT)~\citep{pagnoni2025byte} merges predictable byte spans into latent tokens using entropy-guided segmentation. Similar ideas have recently been explored for time series. EntroPE~\citep{abeywickrama2025entrope} adapts BLT to time series and uses a separately trained entropy model to place boundaries at high-uncertainty time steps, and TimeSqueeze~\citep{ankireddytimesqueeze} relies on local signal statistics %together with a lightweight state-space encoder 
to adjust patch size based on abrupt changes and local variance. However, these approaches depend on modality-specific heuristics, identifiers, or auxiliary models, which prevents the patching strategy from being jointly optimized with the main backbone and may limit its generalization across domains.

\textbf{End-to-end patching.}
Beyond approaches that rely on external identifiers, handcrafted statistics, or auxiliary models, a more direct alternative is to learn the patching strategy jointly with the downstream objective. In this line of work, Nawrot et al.~\citep{nawrot2023efficient} introduced a trainable boundary predictor that dynamically pools tokens, but the method suffers from training instabilities that have thus far limited its scalability to larger models and deeper hierarchical settings. More recently, H-Net~\citep{hwang2025dynamic}, inspired by U-Net~\citep{ronneberger2015u}, proposed a hierarchical compression architecture that processes sequences at multiple resolutions and uses a state-space model to improve the efficiency of byte-level modeling. Its smoothed routing mechanism enables gradient-based end-to-end patch learning, but because boundary decisions depend on the hidden representation manifold and its evolving dynamics, the patching strategy remains tightly coupled to end-to-end training. This coupling introduces several limitations. First, it prevents a train-short \& continuous-train-long regime, which is undesirable when one would like to pre-train a patching module once and freeze its weights during continued pre-training or fine-tuning of the main model. Second, it requires at least one gradient-carrying backbone block at the lowest resolution within the main model. In settings with extremely fine-grained base units, such as character- or byte-level language modeling, this can remain a significant computational bottleneck. H-Net mitigates this issue by using a state-space model at the lowest resolution, but this choice also narrows its applicability, making the approach suited to causal modeling but not non-causal sequence modeling and masked prediction objectives. Independent and concurrent to our work, \citet{dauncey2026you} investigated the use of reinforcement learning for tokenization in language applications. Aside from different application fields, it differs from ours in optimization algorithm, compression rate enforcement, and hierarchical modeling support. Both H-Net and \citet{dauncey2026you} have a soft compression rate enforcement, and we argue in Section~\ref{sec:proposed-method} and Appendix~\ref{appendix:further_modeling_discussion} that it hinders the training framework from fully reaping the benefits of sequence patching models.

\section{Generative Sequence Patching and Unified MDP Formulation}
\label{appendix:extended_modeling_framework}

Recall that the objective is $\sum_{t=0}^{T-1}\mathcal{L}(q_\phi(x_{0:t}, b_{-1:t}), x_{t+1})$ in generative modeling, and the next point $x_{T+1}$ is generated through $q_\phi(\mathbf{x}_T, \mathbf{b}_T)$ during inference time. Since patching decisions are sequential under this setting, the natural formulation is a sequential decision process as follows.

\paragraph{Generative sequence patching as multi-step MDP} Define state and action at step $t$ as
\begin{align}
    \label{eq:multi-step-mdp-sa}
    & s_t \triangleq \left(x_{0:t}, b_{-1:(t-1)}\right), \quad a_t \triangleq b_t,
\end{align}
The MDP is characterized by the following patching policy, transition dynamics, and reward function:
\begin{align}
    \pi(a_t \mid s_t) & \triangleq \pi_\theta \left(b_t \mid x_{0:t}, b_{-1:(t-1)} \right) \\
    P(s_{t+1} \mid s_t, a_t) & \triangleq p(x_{t+1} \mid x_{0:t}) \cdot \delta_{x_{0:t}} \cdot \delta_{b_{-1:t}} \\
    R(s_t, a_t) & \triangleq -\mathcal{L} \left(q_\phi \left(x_{0:t}, b_{-1:t} \right), x_{t+1} \right).
\end{align}
In training, the sequence transition dynamic $p(x_{t+1} \mid x_{0:t})$ is induced by the data; in inference, sequential generation is performed by approximating $p(x_{t+1} \mid x_{0:t})$ with the learned model $q_\phi \left(x_{0:t}, b_{-1:t} \right)$.

Although conditioning decision $b_t$ on the previously sampled boundary decisions $b_{0:(t-1)}$ could be more expressive, such a strategy is significantly less efficient to train for long context windows, especially with a transformer backbone. Therefore, similar to H-Net \cite{hwang2025dynamic}, we let boundary decisions at steps $t_1$ and $t_2$ be independently sampled for $t_1 \neq t_2$, yielding the simplified factorization below.

\paragraph{Unified one-step MDP for efficient optimization} Let $s_t \triangleq x_{0:t}$ in place of the definition in Eq.~\ref{eq:multi-step-mdp-sa}. The MDP is characterized by
\begin{align}
    \pi(a_t \mid s_t) &\triangleq \pi_\theta \left(b_t \mid x_{0:t} \right) \label{eq:pi-in-one-step-mdp}\\
    P(s_{t+1} \mid s_t, a_t) & \triangleq p(x_{t+1} \mid x_{0:t}) \cdot \delta_{x_{0:t}} \\
    R(s_t, a_t) &\triangleq \begin{cases}
        -\sum_{i=0}^t \mathcal{L} \left(q_\phi \left(x_{0:i}, b_{-1:i} \right), x_{i+1} \right), & \text{if } t = T \\
        0, & \text{otherwise}.
    \end{cases}
\end{align}
Although the formulation itself still appears to be a multi-step MDP, this modification enables parallelized decision making and reward computation during training and makes the MDP a realization of the contextual bandit in Eq.~\ref{eq:contextual-bandit}, since the trajectory
\begin{align}
    \tau(\theta) = P(s_0) \prod_{t=0}^T \pi(a_t \mid s_t) P(s_{t+1} \mid s_t, a_t) = p(\mathbf{x}_T) \prod_{t=0}^T \pi_\theta (b_t \mid x_{0:t})
\end{align}
can be computed in one forward pass of $\pi_\theta$, and the terminal reward can then be assessed in one forward pass of $q_\phi$. We use this unified setup as our formulation, where causality can be enforced if desired so long as the models satisfy
\begin{align}
    q_\phi \left(\mathbf{x}_T, \mathbf{b}_T \right)_{[t]} = q_\phi \left(x_{0:t}, b_{-1:t} \right), \quad \pi_\theta \left( b_{0:t} \mid \mathbf{x}_T \right)_{[t]} = \pi_\theta \left( b_t \mid x_{0:t} \right),
\end{align}
where $M_{[t]}$ is the $t$-th column vector (or element) of matrix (or vector) $M$. This is automatically the case for standard causal sequence backbones such as uni-directional SSMs and transformers with causal masks. During inference, one can compute the right-side boundary logits of each time step sequentially, as in Eq.~\ref{eq:pi-in-one-step-mdp}, on tasks where transition dynamic $p(x_{t+1} \mid x_{0:t})$ needs to be approximated by sequence model $q_\phi \left(x_{0:t}, b_{-1:t} \right)$.

Note that the one-step MDP view is a \emph{training-time reward-assignment scheme}, not a claim that boundary decisions have to be made non-sequentially: the sharing of one terminal reward across actions is what makes the group-relative advantage well-defined. The same formulation reduction applies to GRPO for sequential language-model generation, whose outcome-supervised objective is likewise equivalent, in its reward assignment, to a contextual bandit in which the entire sampled text output can be viewed as a single action receiving one terminal reward. A causal \model therefore follows by masking the policy to attend only to $x_{\le t}$, which leaves the reward, advantage, and GRPG update unchanged and emits boundaries in a single pass at inference, similar to H-Net~\cite{hwang2025dynamic}.

\section{Training Algorithm}
\label{appendix:training_algorithm}

An algorithm block for the training workflow of our approach is presented in Algorithm~\ref{alg:reinpatch_training}.

\begin{algorithm}[tb]
\caption{Joint End-to-End Training of \model and Downstream Backbone}
\label{alg:reinpatch_training}
\begin{algorithmic}[1]
\Require Dataset $\mathcal{D}$, initialized patching policy $\pi_\theta$, initialized downstream sequence model $q_\phi$
\Require Group size $G$, minimum compression rate $r_m$, learning rates $\eta_\theta, \eta_\phi$
\While{not converged}
    \State Sample a mini-batch of sequences $\mathbf{x}_T$ and targets $y$ from $\mathcal{D}$
    \For{each sequence in the mini-batch}
        \For{$i = 1, \dots, G$}
            \State \textcolor{orange}{Sample boundary decisions $b_{0:T}^{(i)} \sim \pi_\theta(\cdot \mid \mathbf{x}_T)$} \Comment{\textcolor{orange}{The policy for $q_\phi$}}
            \If{\textcolor{teal}{compression rate $(T+1) / \sum_{t=0}^T b_t^{(i)} < r_m$}} \Comment{\textcolor{teal}{The environment for $\pi_\theta$}}
                \State \textcolor{teal}{Merge extra patches from the right side of $\mathbf{b}_T^{(i)}$}
            \EndIf
            \State \textcolor{teal}{Compute backbone prediction $\hat{y}^{(i)} = q_\phi(\mathbf{x}_T, \mathbf{b}_T^{(i)})$}
            \State \textcolor{teal}{Compute task loss $\mathcal{L}^{(i)} = \mathcal{L}(\hat{y}^{(i)}, y)$}
            \State \textcolor{teal}{Set terminal reward $R^{(i)} = -\mathcal{L}^{(i)}$}
        \EndFor
        \State Calculate group reward statistics: 
        \Statex \qquad \qquad $\mu_R = \frac{1}{G} \sum_{i=1}^G R^{(i)}$ \quad and \quad $\sigma_R = \sqrt{\frac{1}{G} \sum_{i=1}^G (R^{(i)} - \mu_R)^2}$
        \For{$i = 1, \dots, G$}
            \State Compute advantage $A^{(i)} = \frac{R^{(i)} - \mu_R}{\sigma_R + \epsilon}$
        \EndFor
        \State \textcolor{brown}{Compute policy gradient (GRPG):} \Comment{\textcolor{brown}{For policy $\pi_\theta$ update}}
        \Statex \qquad \qquad \textcolor{brown}{$g_\theta = \frac{1}{G} \sum_{i=1}^G \nabla_\theta \log \pi_\theta(b_{0:T}^{(i)} \mid \mathbf{x}_T) A^{(i)}$}
        \State \textcolor{olive}{Compute backbone gradient:} \Comment{\textcolor{olive}{For environment $q_\phi$ update}}
        \Statex \qquad \qquad \textcolor{olive}{$g_\phi = \nabla_\phi \frac{1}{G} \sum_{i=1}^G \mathcal{L}^{(i)}$}
    \EndFor
    \State Average gradients $g_\theta$ and $g_\phi$ across the mini-batch
    \State Update policy network: $\theta \leftarrow \theta + \eta_\theta g_\theta$
    \State Update backbone network: $\phi \leftarrow \phi - \eta_\phi g_\phi$
\EndWhile
\end{algorithmic}
\end{algorithm}

\section{Further Modeling Discussion}
\label{appendix:further_modeling_discussion}

\paragraph{Why is GRPO not preferred over GRPG?} The design of group mean as critic instead of neural network as critic is most recently well-known under the LLM mathematical reasoning line of work as group relative policy optimization (GRPO) \cite{shao2024deepseekmath}:
\begin{align}
    \label{eq:grpo}
    \mathcal{J}_{GRPO} = \frac{1}{G} \sum_{i=1}^G \mathbb E \left[ \sum_{t=0}^T \min \left\{ r_\theta A^{(i)}, \mathrm{clip} \left\{ r_\theta A^{(i)}, 1-\delta, 1+\delta \right\} \right\} \right] - \beta D_{KL}\left[ \pi_\theta \parallel \pi_{ref} \right]
\end{align}
where $r_\theta = \frac{\pi_\theta(a_t^{(i)} \mid s_t^{(i)})}{\pi_{\theta_{old}}(a_t^{(i)} \mid s_t^{(i)})}$ and $A^{(i)} = \frac{R(s_T^{(i)}, a_T^{(i)}) - \mu_R}{\sigma_R + \epsilon}$. On the other hand, The GRPG objective in Eq.~\ref{eq:grpg} can be written as
\begin{align}
    \label{eq:grpg-alt}
    \nabla_\theta \mathcal{J}_{GRPG} = \frac{1}{G} \sum_{i=1}^G \mathbb E \left[ \sum_{t=0}^T \nabla_\theta \log \pi_\theta (a_t^{(i)} \mid s_t^{(i)}) \frac{R(s_T^{(i)}, a_T^{(i)}) - \mu_R}{\sigma_R + \epsilon}\right]
\end{align}
if we let $a_t \triangleq b_t$ and $s_t \triangleq x_{0:T}$ (or $s_t \triangleq x_{0:t}$ in the unified one-step MDP setup for the causal setting described in Appendix~\ref{appendix:extended_modeling_framework}). Despite having the same advantage function, Eq.~\ref{eq:grpo} has two distinct components compared to Eq.~\ref{eq:grpg-alt}: 1) the PPO~\cite{schulman2017proximal} objective in place of policy gradient for multi-epoch mini-batch update, 2) reference policy $\pi_{ref}$ as anchor.

For the first part, the reason single-epoch updates are safer (hence, policy gradient is preferred over PPO) is that the environment in our case is a continuously evolving and highly parallelizable neural model $q_\theta$. Algorithms like PPO do multi-epoch mini-batch updates to squeeze as much value as possible out of a batch of rollouts because interacting with their environments (e.g. physics simulators or math verifiers) is slow and expensive. In our setup, sample generation is cheap. Doing a single update step on fresh data (i.e. pure on-policy learning) prevents the policy from taking overly large, destructive steps and avoids the need for complex clipping mechanisms.

For the second part, the answer is simple: this work focuses on the from-scratch training and pre-training of sequence patching policy, where we do not have a pre-trained reference policy and we do not explore fine-tuning patching policy under the scope of this paper. We have an online learning setup where the reward function fully capture the intended goal of the patching policy -- to find the patching boundaries that result in the best sequence model performance. Besides, contrary to conversational AI such as LLMs, we do not require diversity in its generation. It is not a problem if the optimized patcher always provides the same patching strategy for a given sequence. If early exploration is of particular concern, practitioners are free to apply standard exploration strategies as well to prevent premature convergence, such as adding entropy term $\mathcal{H}(\pi_\theta)$ with coefficient $\beta$ to the objective function instead of $\beta D_{KL}\left[ \pi_\theta \parallel \pi_{ref} \right]$.

\paragraph{How is desired compression rate enforced?} In prior work such as \citet{hwang2025dynamic}, a target compression rate is encouraged by an auxiliary loss $\mathcal{L}_{aux}$ that increases when the compression rate under the sampled patching boundaries deviates from the target compression rate. This prevents model from converging to a trivial solution, such as placing patching boundaries everywhere, or not placing boundaries at all (since the backbone model has residual connection). On the contrary, the objective of our patching strategy is solely to optimize the performance of the downstream sequence model $q_\theta$, so how can such reward hacking be prevented?

To answer this question, we first argue that it is systematically better to set desired compression rate as a strict limit rather than as a soft auxiliary objective as in H-Net. The reason is its advantage in model size pre-planning. Soft-encouragement of compression rate means practitioners always need to pre-plan model dimensions for the worst case scenario, which is having no compression (i.e. model places patching boundaries everywhere) in the beginning of training, or at any point during training. However, one of the key benefits of patching is the scalability: it allows a deeper neural model with extended hidden size as the spatial or temporal horizons concentrate, and H-Net is not able to fully reap such benefit. After all, if the algorithm truly thinks that setting boundaries everywhere is going to yield the best performance, then it should be allowed to do so. The real problem here is that it cannot enlarge model size during the course of training to adapt to the shrinking context length to fully understand the benefit of doing compression.

On the contrary, setting strict minimum compression rate can be achieved under our proposed method, since we do not rely on gradient flow from the main model to update the patcher. We enforce the minimum compression rate $r_m$ in $q_\phi$, in a way such that if $\pi_\theta$ proposes a set of boundaries that yields a lower compression rate than $r_m$, all extra patches on the right side are merged into one patch. In other words, the information regarding the preset compression rate requirement is in the \textit{environment} of $\pi_\theta$, and $\pi_\theta$ has to figure it out and adapt to it during the course of training. Note that we perform merging on the right side to make the environment more challenging for $\pi_\theta$, since recent information are usually perceived to be more meaningful for sequence data. With this strict enforcement, we need not pre-plan model sizes for the worst case scenario, and can reap the full benefits of deep patching models.

\section{Sequence Model Architecture}
\label{appendix:model_architecture}

\begin{figure}
    \centering
    \includegraphics[width=0.88\linewidth]{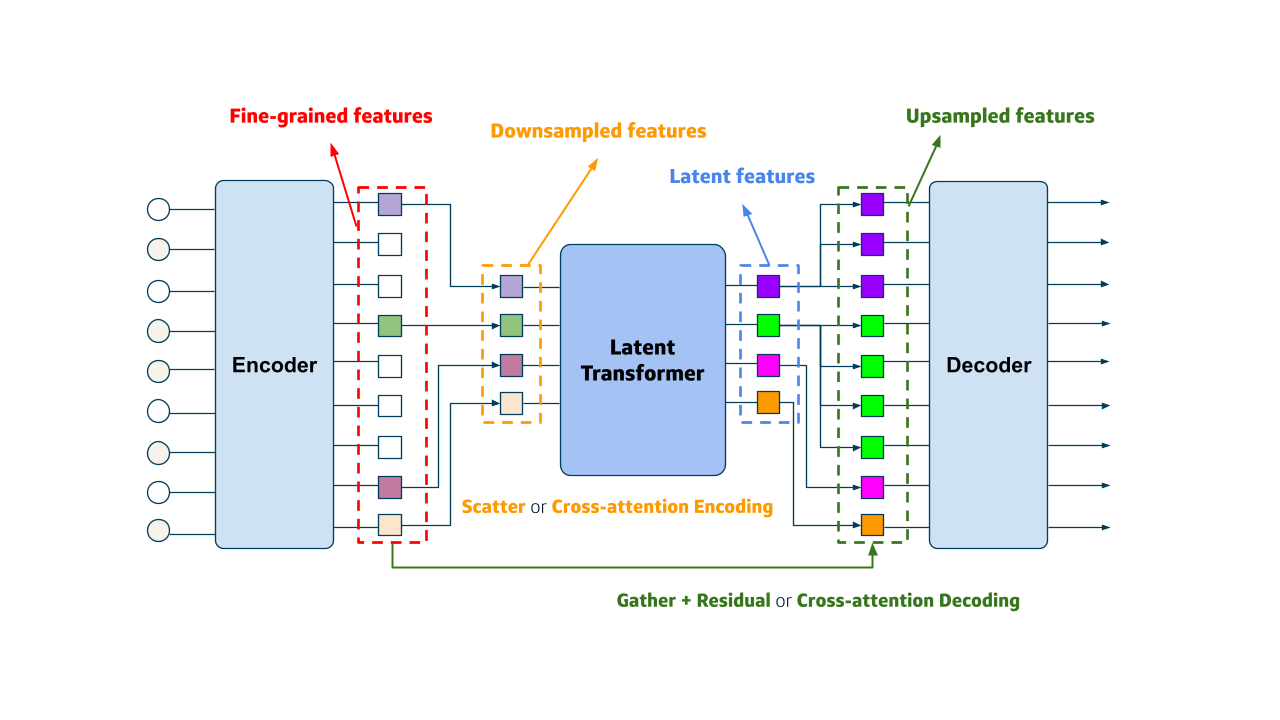}
    \caption{Sequence backbone model. Note that this illustration is for contextual tasks. For generative (i.e. causal) tasks, patches are right-shifted by one (and a special token embedding is added to the beginning of the sequence) before upsampling.}
    \label{fig:model_arch}
\end{figure}

We use a standard sequence patching backbone model widely used in prior work, illustrated in Figure~\ref{fig:model_arch}. For details on cross-attention encoding and decoding, see \citet{pagnoni2025byte}; for details on the scatter and gather operations, see \citet{hwang2025dynamic}. This design extends to hierarchical patching when multiple levels of patching boundaries are provided by stacking encoders and decoders on downsampled and upsampled features, analogous to how multi-resolution architectures such as U-Net stack operations across resolution levels. We use a two-level hierarchical model for our end-to-end patching method in the main experiments.

For downsampling, we share the view of \citet{hwang2025dynamic} that cross-attention encoding in \citet{pagnoni2025byte} is not always necessary, but we maintain that cross-attention decoding can be beneficial. The gather upsampling approach, which repeats the same hidden state for every point in a patch, is less compatible with down-projection and can introduce information loss. In U-Net-style architectures, the consecutive operations of up-projection and downsampling, as well as upsampling and down-projection, allow information to transfer between spatial and hidden dimensions to mitigate compression loss. This does not hold for the combination of gather upsampling and down-projection, since information compressed in hidden dimensions cannot transfer back to temporal dimensions. This reduces the benefit of hierarchical modeling and increases reliance on the residual connection.

\section{Experiment Details}
\label{appendix:experiment-details}

\subsection{Datasets}
\label{appendix:datasets}

Table~\ref{tab:dataset_stats} summarizes the statistics of the datasets used in our forecasting evaluation, including their domain, sampling frequency, length, dimensionality, and train/validation/test split.

\begin{table*}[!ht]
\centering
\small
\caption{Statistics of datasets.}
\label{tab:dataset_stats}
\setlength{\tabcolsep}{6pt}
\renewcommand{\arraystretch}{1.12}
\resizebox{0.98\textwidth}{!}{%
\begin{tabular}{llccrrl}
\toprule
\rowcolor{gray!12}
Dataset & Domain & Frequency & Lengths & Dim & Split & Description \\
\midrule
ETTh1 & Electricity & 1 hour & 14,400 & 7 & 6:2:2 & Power transformer 1, comprising seven indicators such as oil temperature and useful load \\
ETTh2 & Electricity & 1 hour & 14,400 & 7 & 6:2:2 & Power transformer 2, comprising seven indicators such as oil temperature and useful load \\
ETTm1 & Electricity & 15 mins & 57,600 & 7 & 6:2:2 & Power transformer 1, comprising seven indicators such as oil temperature and useful load \\
ETTm2 & Electricity & 15 mins & 57,600 & 7 & 6:2:2 & Power transformer 2, comprising seven indicators such as oil temperature and useful load \\
Weather & Environment & 10 mins & 52,696 & 21 & 7:1:2 & Recorded every day for the whole year 2020, which contains 21 meteorological indicators \\
Electricity & Electricity & 1 hour & 26,304 & 321 & 7:1:2 & Electricity records the electricity consumption in kWh every 1 hour from 2012 to 2014 \\
Solar & Energy & 10 mins & 52,560 & 137 & 7:1:2 & Solar production records collected from 137 PV plants in Alabama \\
Traffic & Traffic & 1 hour & 17,544 & 862 & 7:1:2 & Road occupancy rates measured by 862 sensors on San Francisco Bay area freeways \\
\bottomrule
\end{tabular}
}
\end{table*}

\subsection{Problem Formulation}
\label{appendix:problem-formulation}

Consider a multivariate time series $\mathbf{x}_T = x_{0:T}$ with $C$ channels, yielding an input $\mathbf{x}_T \in \mathbb R^{C \times T}$. The forecasting model takes the most recent $T$ time steps from each channel as input and predicts the next $H$ time steps for all channels. Following~\cite{nie2023patch}, we adopt the channel-independence principle and process the multivariate input as a collection of univariate series with shared model parameters, then aggregate the per-channel predictions back into a multivariate forecast. This is the standard paradigm established by recent forecasting architectures \cite{nie2023patch, zeng2023transformers, wuenhancing}.

\subsection{Training and Selection Details}
\label{appendix:training-selection-details}

For all forecasting experiments, we use the official train/validation/test splits associated with each benchmark dataset and select hyperparameters by validation MSE. Candidate look-back lengths are $T \in \{96,192,336\}$, and all tables report the test result from the validation-selected checkpoint and configuration. Unless otherwise stated, runs use random seeds $\{1,2,3,4,5\}$; reported means are seed averages and reported $\pm$ values are standard deviations across seeds.

The downstream forecasting backbone is optimized with Adam and MSE loss. Forecasting runs use OneCycleLR, early stopping on validation loss, batch sizes selected from $\{32,64,128,256\}$ depending on the dataset and horizon, and training budgets between $20$ and $80$ epochs with patience between $7$ and $20$. All forecasting experiments of \model are conducted using PyTorch~\citep{paszke2019pytorch} in Python 3.11 and executed on an NVIDIA A100 GPU.
% The \model end-to-end patcher is a lightweight transformer with three layers, hidden size $32$, dropout $0.1$, and attention heads selected from the validation grid. 
% Group-relative policy updates use group size $G \in \{2,3\}$ and a patcher learning rate of $10^{-2}$ in the end-to-end forecasting runs.

For the zero-shot patching experiments, the foundation patcher is jointly pre-trained with a forecasting backbone on UTSD using sequence length $240$, prediction length $96$, two epochs, batch size $1024$, Adam learning rate $10^{-4}$ for the backbone, patcher learning rate $3{\times}10^{-3}$, group size $3$, and minimum compression rate $8$. The pre-trained patcher uses four transformer layers, hidden size $32$, two attention heads, and dropout $0.1$. Pre-training are run across two distributed nodes, each with eight NVIDIA A100 GPUs. During downstream zero-shot evaluation, the patcher weights are frozen and only the task-specific forecasting backbone is trained.

% Note: remove for pre-print and camera-ready
Our code is available in the supplementary material.

\subsection{Encoder-Decoder configuration}
\label{appendix:encoder-decoder}

\begin{table}%[!htbp]
\caption{Architecture components of different forecasting methods.}
\label{tab:arch_config}
\centering
\fontsize{7}{8}\selectfont
\renewcommand{\arraystretch}{1.0}
\setlength{\tabcolsep}{6pt}
\begin{tabular}{@{}lccc@{}}
\toprule
\textbf{Method} & \textbf{Encoder ($\times$levels)} & \textbf{Decoder ($\times$levels)} & \textbf{Latent Model} \\
\midrule
Static Patching &  Transformer $\times 1$&  Transformer $\times 1$&  Transformer\\
Selective Patching &  Transformer $\times 1$&  Transformer $\times 1$&  Transformer\\
Periodic Patching &  Transformer $\times 1$&  Transformer $\times 1$&  Transformer\\
Entropy Patching &  Transformer $\times 1$&  Transformer $\times 1$&  Transformer\\
TimeSqueeze's Compression &  SSM $\times 1$&  SSM $\times 1$&  Transformer\\
H-Net's Chunking &  SSM $\times 2$&  SSM $\times 2$&  Transformer\\
\midrule
\model\ (end-to-end) [Ours] &  Transformer $\times 2$&  Transformer $\times 2$&  Transformer\\
\model\ (zero-shot) [Ours] &  Transformer $\times 1$&  Transformer $\times 1$&  Transformer\\
\bottomrule
\end{tabular}
\end{table}

Based on the baseline chosen, patching is implemented via either a cross-attention-based transformer or an SSM. For TimeSqueeze's compression and H-Net's chunking, we use an SSM backbone for the downsampling encoder and upsampling decoder, aligned with the original literature \cite{ankireddytimesqueeze, hwang2025dynamic}. For the remaining patching methods, we use a transformer backbone. Exact configurations for each method are listed in Table~\ref{tab:arch_config}. We choose only one level of compression for the \model foundation zero-shot patcher to keep it transferable and compatible with forecasting methods that may not support multi-level hierarchical patching. H-Net and \model are the only evaluated methods capable of supporting multi-level hierarchical patching.

\subsection{Full Results}
\label{appendix:additional_results}

While Table~\ref{tab:results_avg} in the main text reports MSE and MAE averaged over the four forecasting horizons, Table~\ref{tab:results_full} presents the complete per-horizon breakdown ($H\in\{96,192,336,720\}$) for all eight methods across the eight datasets, with each cell averaged over five seeds. The finer granularity confirms the trend seen in the averaged results: \model (zero-shot and end-to-end) takes the majority of first-place entries and stays competitive at both short and long horizons, indicating that its averaged advantage does not stem from a few favorable horizons but holds consistently as the forecasting difficulty grows.

\begin{table*}[!htbp]
\caption{Multivariate forecasting full results with forecasting horizons $H \in \{96, 192, 336, 720\}$ for the datasets. \textcolor{red}{\textbf{Red}}: the best, \textcolor{blue}{\underline{Blue}}: the 2nd best.}
\label{tab:results_full}
\centering
\scriptsize
\renewcommand{\arraystretch}{1.0}
\setlength{\tabcolsep}{4pt}
\resizebox{\textwidth}{!}{
    \begin{tabular}{@{}cc|cc|cc|cc|cc|cc|cc|cc|cc@{}}
    \toprule
        \multicolumn{2}{c|}{\textbf{Models}} & \multicolumn{2}{c|}{\shortstack[c]{\textbf{ReinPatch (zero-shot)}\\\textbf{[Ours]}}} & \multicolumn{2}{c|}{\shortstack[c]{\textbf{ReinPatch (end-to-end)}\\\textbf{[Ours]}}} & \multicolumn{2}{c|}{\shortstack[c]{\textbf{Static Patching}\\\textbf{\cite{nie2023patch}}}} & \multicolumn{2}{c|}{\shortstack[c]{\textbf{Selective Patching}\\\textbf{\cite{wuenhancing}}}} & \multicolumn{2}{c|}{\shortstack[c]{\textbf{Periodic Patching}\\\textbf{\cite{wang2025lightgts}}}} & \multicolumn{2}{c|}{\shortstack[c]{\textbf{Entropy Patching}\\\textbf{\cite{pagnoni2025byte,abeywickrama2025entrope}}}} & \multicolumn{2}{c|}{\shortstack[c]{\textbf{TimeSqueeze's}\\\textbf{Compression~\cite{ankireddytimesqueeze}}}} & \multicolumn{2}{c}{\shortstack[c]{\textbf{H-Net's Chunking}\\\textbf{\cite{hwang2025dynamic}}}} \\
        \multicolumn{1}{c|}{\textbf{Metrics}} & \multicolumn{1}{c|}{\textbf{F.H.}} & \multicolumn{1}{@{\hskip\tabcolsep}c@{\hskip0.5\tabcolsep}}{\textbf{MSE}} & \multicolumn{1}{@{\hskip0.5\tabcolsep}c@{\hskip\tabcolsep}|}{\textbf{MAE}} & \multicolumn{1}{@{\hskip\tabcolsep}c@{\hskip0.5\tabcolsep}}{\textbf{MSE}} & \multicolumn{1}{@{\hskip0.5\tabcolsep}c@{\hskip\tabcolsep}|}{\textbf{MAE}} & \multicolumn{1}{@{\hskip\tabcolsep}c@{\hskip0.5\tabcolsep}}{\textbf{MSE}} & \multicolumn{1}{@{\hskip0.5\tabcolsep}c@{\hskip\tabcolsep}|}{\textbf{MAE}} & \multicolumn{1}{@{\hskip\tabcolsep}c@{\hskip0.5\tabcolsep}}{\textbf{MSE}} & \multicolumn{1}{@{\hskip0.5\tabcolsep}c@{\hskip\tabcolsep}|}{\textbf{MAE}} & \multicolumn{1}{@{\hskip\tabcolsep}c@{\hskip0.5\tabcolsep}}{\textbf{MSE}} & \multicolumn{1}{@{\hskip0.5\tabcolsep}c@{\hskip\tabcolsep}|}{\textbf{MAE}} & \multicolumn{1}{@{\hskip\tabcolsep}c@{\hskip0.5\tabcolsep}}{\textbf{MSE}} & \multicolumn{1}{@{\hskip0.5\tabcolsep}c@{\hskip\tabcolsep}|}{\textbf{MAE}} & \multicolumn{1}{@{\hskip\tabcolsep}c@{\hskip0.5\tabcolsep}}{\textbf{MSE}} & \multicolumn{1}{@{\hskip0.5\tabcolsep}c@{\hskip\tabcolsep}|}{\textbf{MAE}} & \multicolumn{1}{@{\hskip\tabcolsep}c@{\hskip0.5\tabcolsep}}{\textbf{MSE}} & \multicolumn{1}{@{\hskip0.5\tabcolsep}c@{\hskip\tabcolsep}}{\textbf{MAE}} \\
        \midrule
        \multirow{4}{*}{\raisebox{10pt}{\rotatebox[origin=l]{90}{ETTh1}}} & 96 & \textcolor{red}{\textbf{0.360}} $\pm$ 0.008 & \textcolor{red}{\textbf{0.400}} $\pm$ 0.006 & 0.374 $\pm$ 0.007 & 0.405 $\pm$ 0.006 & 0.364 & 0.402 & \textcolor{blue}{\underline{0.363}} & 0.401 & 0.366 & 0.404 & 0.378 & 0.401 & 0.367 & \textcolor{blue}{\underline{0.400}} & 0.384 & 0.401 \\
         & 192 & \textcolor{red}{\textbf{0.394}} $\pm$ 0.003 & \textcolor{red}{\textbf{0.419}} $\pm$ 0.004 & \textcolor{blue}{\underline{0.410}} $\pm$ 0.016 & \textcolor{blue}{\underline{0.423}} $\pm$ 0.009 & 0.415 & 0.427 & 0.417 & 0.428 & 0.419 & 0.431 & 0.441 & 0.440 & 0.429 & 0.425 & 0.436 & 0.432 \\
         & 336 & \textcolor{red}{\textbf{0.395}} $\pm$ 0.007 & \textcolor{blue}{\underline{0.423}} $\pm$ 0.005 & 0.397 $\pm$ 0.007 & 0.424 $\pm$ 0.006 & 0.402 & 0.427 & 0.404 & 0.427 & \textcolor{blue}{\underline{0.395}} & \textcolor{red}{\textbf{0.421}} & 0.424 & 0.445 & 0.420 & 0.439 & 0.438 & 0.434 \\
         & 720 & \textcolor{red}{\textbf{0.440}} $\pm$ 0.019 & \textcolor{red}{\textbf{0.455}} $\pm$ 0.010 & \textcolor{blue}{\underline{0.441}} $\pm$ 0.016 & \textcolor{blue}{\underline{0.461}} $\pm$ 0.011 & 0.453 & 0.463 & 0.447 & 0.462 & 0.471 & 0.478 & 0.457 & 0.465 & 0.459 & 0.468 & 0.478 & 0.479 \\
        \midrule
        \multirow{4}{*}{\raisebox{10pt}{\rotatebox[origin=l]{90}{ETTh2}}} & 96 & 0.281 $\pm$ 0.006 & 0.339 $\pm$ 0.004 & 0.271 $\pm$ 0.005 & 0.338 $\pm$ 0.006 & \textcolor{red}{\textbf{0.265}} & \textcolor{red}{\textbf{0.328}} & \textcolor{blue}{\underline{0.268}} & \textcolor{blue}{\underline{0.334}} & 0.288 & 0.345 & 0.283 & 0.349 & 0.270 & 0.338 & 0.284 & 0.341 \\
         & 192 & 0.334 $\pm$ 0.007 & 0.380 $\pm$ 0.005 & 0.330 $\pm$ 0.010 & \textcolor{red}{\textbf{0.374}} $\pm$ 0.007 & 0.335 & 0.380 & \textcolor{blue}{\underline{0.328}} & \textcolor{blue}{\underline{0.376}} & 0.332 & 0.378 & 0.335 & 0.381 & \textcolor{red}{\textbf{0.328}} & 0.377 & 0.337 & 0.385 \\
         & 336 & \textcolor{red}{\textbf{0.314}} $\pm$ 0.006 & \textcolor{red}{\textbf{0.374}} $\pm$ 0.004 & 0.322 $\pm$ 0.009 & 0.380 $\pm$ 0.010 & 0.331 & 0.385 & \textcolor{blue}{\underline{0.320}} & \textcolor{blue}{\underline{0.378}} & 0.326 & 0.386 & 0.329 & 0.383 & 0.325 & 0.383 & 0.327 & 0.384 \\
         & 720 & \textcolor{red}{\textbf{0.362}} $\pm$ 0.008 & \textcolor{red}{\textbf{0.415}} $\pm$ 0.006 & \textcolor{blue}{\underline{0.367}} $\pm$ 0.008 & \textcolor{blue}{\underline{0.418}} $\pm$ 0.007 & 0.409 & 0.439 & 0.405 & 0.444 & 0.398 & 0.428 & 0.410 & 0.442 & 0.412 & 0.442 & 0.410 & 0.440 \\
        \midrule
        \multirow{4}{*}{\raisebox{10pt}{\rotatebox[origin=l]{90}{ETTm1}}} & 96 & \textcolor{blue}{\underline{0.292}} $\pm$ 0.004 & \textcolor{blue}{\underline{0.346}} $\pm$ 0.003 & \textcolor{red}{\textbf{0.291}} $\pm$ 0.005 & \textcolor{red}{\textbf{0.345}} $\pm$ 0.003 & 0.293 & 0.347 & 0.302 & 0.354 & 0.307 & 0.355 & 0.310 & 0.353 & 0.311 & 0.354 & 0.299 & 0.347 \\
         & 192 & \textcolor{blue}{\underline{0.330}} $\pm$ 0.004 & 0.371 $\pm$ 0.002 & \textcolor{red}{\textbf{0.329}} $\pm$ 0.003 & \textcolor{red}{\textbf{0.370}} $\pm$ 0.003 & 0.334 & 0.376 & 0.351 & 0.388 & 0.335 & 0.375 & 0.335 & 0.375 & 0.338 & \textcolor{blue}{\underline{0.371}} & 0.343 & 0.371 \\
         & 336 & \textcolor{red}{\textbf{0.357}} $\pm$ 0.002 & \textcolor{red}{\textbf{0.389}} $\pm$ 0.001 & \textcolor{blue}{\underline{0.357}} $\pm$ 0.003 & \textcolor{blue}{\underline{0.389}} $\pm$ 0.002 & 0.371 & 0.398 & 0.371 & 0.401 & 0.370 & 0.397 & 0.369 & 0.393 & 0.371 & 0.392 & 0.381 & 0.394 \\
         & 720 & \textcolor{blue}{\underline{0.409}} $\pm$ 0.003 & \textcolor{blue}{\underline{0.422}} $\pm$ 0.002 & 0.410 $\pm$ 0.002 & 0.423 $\pm$ 0.001 & \textcolor{red}{\textbf{0.409}} & \textcolor{red}{\textbf{0.420}} & 0.413 & 0.423 & 0.414 & 0.426 & 0.418 & 0.423 & 0.428 & 0.433 & 0.427 & 0.423 \\
        \midrule
        \multirow{4}{*}{\raisebox{10pt}{\rotatebox[origin=l]{90}{ETTm2}}} & 96 & \textcolor{red}{\textbf{0.168}} $\pm$ 0.002 & \textcolor{red}{\textbf{0.257}} $\pm$ 0.003 & \textcolor{blue}{\underline{0.168}} $\pm$ 0.002 & \textcolor{blue}{\underline{0.257}} $\pm$ 0.002 & 0.174 & 0.261 & 0.172 & 0.261 & 0.172 & 0.259 & 0.176 & 0.263 & 0.175 & 0.265 & 0.177 & 0.265 \\
         & 192 & \textcolor{red}{\textbf{0.225}} $\pm$ 0.004 & \textcolor{red}{\textbf{0.297}} $\pm$ 0.003 & 0.228 $\pm$ 0.007 & 0.299 $\pm$ 0.004 & 0.232 & 0.306 & \textcolor{blue}{\underline{0.227}} & \textcolor{blue}{\underline{0.297}} & 0.235 & 0.302 & 0.244 & 0.310 & 0.232 & 0.308 & 0.229 & 0.300 \\
         & 336 & \textcolor{red}{\textbf{0.281}} $\pm$ 0.007 & \textcolor{red}{\textbf{0.335}} $\pm$ 0.006 & \textcolor{blue}{\underline{0.283}} $\pm$ 0.008 & \textcolor{blue}{\underline{0.335}} $\pm$ 0.003 & 0.300 & 0.349 & 0.296 & 0.345 & 0.286 & 0.337 & 0.298 & 0.346 & 0.291 & 0.339 & 0.299 & 0.340 \\
         & 720 & \textcolor{blue}{\underline{0.375}} $\pm$ 0.007 & 0.394 $\pm$ 0.005 & \textcolor{red}{\textbf{0.374}} $\pm$ 0.004 & \textcolor{blue}{\underline{0.391}} $\pm$ 0.003 & 0.380 & \textcolor{red}{\textbf{0.388}} & 0.379 & 0.394 & 0.394 & 0.404 & 0.408 & 0.416 & 0.386 & 0.398 & 0.382 & 0.404 \\
        \midrule
        \multirow{4}{*}{\raisebox{10pt}{\rotatebox[origin=l]{90}{Weather}}} & 96 & \textcolor{red}{\textbf{0.149}} $\pm$ 0.002 & 0.203 $\pm$ 0.003 & 0.153 $\pm$ 0.003 & 0.206 $\pm$ 0.002 & 0.155 & 0.205 & 0.155 & 0.207 & 0.153 & 0.206 & 0.157 & 0.205 & \textcolor{blue}{\underline{0.151}} & \textcolor{blue}{\underline{0.202}} & 0.151 & \textcolor{red}{\textbf{0.201}} \\
         & 192 & \textcolor{red}{\textbf{0.191}} $\pm$ 0.001 & \textcolor{red}{\textbf{0.242}} $\pm$ 0.002 & 0.197 $\pm$ 0.003 & 0.246 $\pm$ 0.004 & \textcolor{blue}{\underline{0.195}} & 0.243 & 0.197 & 0.246 & 0.196 & 0.244 & 0.198 & 0.246 & 0.197 & 0.248 & 0.196 & \textcolor{blue}{\underline{0.242}} \\
         & 336 & 0.245 $\pm$ 0.002 & \textcolor{blue}{\underline{0.282}} $\pm$ 0.003 & \textcolor{red}{\textbf{0.244}} $\pm$ 0.001 & 0.282 $\pm$ 0.001 & 0.250 & 0.285 & 0.246 & 0.283 & 0.248 & 0.286 & 0.247 & 0.284 & \textcolor{blue}{\underline{0.244}} & 0.286 & 0.246 & \textcolor{red}{\textbf{0.281}} \\
         & 720 & 0.319 $\pm$ 0.003 & 0.334 $\pm$ 0.003 & 0.321 $\pm$ 0.002 & 0.336 $\pm$ 0.002 & 0.319 & 0.335 & 0.323 & 0.338 & \textcolor{blue}{\underline{0.316}} & \textcolor{red}{\textbf{0.330}} & \textcolor{red}{\textbf{0.316}} & \textcolor{blue}{\underline{0.331}} & 0.323 & 0.341 & 0.324 & 0.335 \\
        \midrule
        \multirow{4}{*}{\raisebox{10pt}{\rotatebox[origin=l]{90}{Electricity}}} & 96 & \textcolor{red}{\textbf{0.137}} $\pm$ 0.003 & \textcolor{red}{\textbf{0.238}} $\pm$ 0.005 & 0.140 $\pm$ 0.003 & 0.242 $\pm$ 0.005 & 0.151 & 0.264 & 0.150 & 0.260 & 0.152 & 0.265 & \textcolor{blue}{\underline{0.139}} & \textcolor{blue}{\underline{0.240}} & 0.140 & 0.242 & 0.141 & 0.246 \\
         & 192 & \textcolor{red}{\textbf{0.153}} $\pm$ 0.001 & \textcolor{red}{\textbf{0.252}} $\pm$ 0.003 & 0.157 $\pm$ 0.003 & 0.258 $\pm$ 0.004 & 0.182 & 0.292 & 0.175 & 0.282 & 0.177 & 0.283 & \textcolor{blue}{\underline{0.157}} & \textcolor{blue}{\underline{0.256}} & 0.163 & 0.264 & 0.159 & 0.258 \\
         & 336 & \textcolor{blue}{\underline{0.172}} $\pm$ 0.002 & \textcolor{blue}{\underline{0.273}} $\pm$ 0.003 & 0.173 $\pm$ 0.001 & 0.282 $\pm$ 0.001 & 0.202 & 0.307 & 0.194 & 0.298 & 0.198 & 0.301 & \textcolor{red}{\textbf{0.172}} & \textcolor{red}{\textbf{0.270}} & 0.179 & 0.277 & 0.176 & 0.274 \\
         & 720 & \textcolor{blue}{\underline{0.209}} $\pm$ 0.005 & \textcolor{blue}{\underline{0.304}} $\pm$ 0.005 & 0.217 $\pm$ 0.003 & 0.312 $\pm$ 0.004 & 0.240 & 0.331 & 0.230 & 0.326 & 0.236 & 0.325 & \textcolor{red}{\textbf{0.208}} & \textcolor{red}{\textbf{0.301}} & 0.217 & 0.318 & 0.223 & 0.323 \\
        \midrule
        \multirow{4}{*}{\raisebox{10pt}{\rotatebox[origin=l]{90}{Solar}}} & 96 & \textcolor{red}{\textbf{0.177}} $\pm$ 0.003 & \textcolor{red}{\textbf{0.241}} $\pm$ 0.007 & 0.185 $\pm$ 0.003 & 0.250 $\pm$ 0.007 & 0.192 & 0.248 & 0.181 & 0.251 & 0.193 & 0.249 & \textcolor{blue}{\underline{0.180}} & \textcolor{blue}{\underline{0.247}} & 0.187 & 0.251 & 0.191 & 0.265 \\
         & 192 & \textcolor{red}{\textbf{0.199}} $\pm$ 0.002 & \textcolor{blue}{\underline{0.265}} $\pm$ 0.002 & 0.202 $\pm$ 0.003 & 0.271 $\pm$ 0.005 & 0.207 & 0.277 & 0.231 & 0.286 & 0.205 & 0.271 & 0.202 & \textcolor{red}{\textbf{0.251}} & \textcolor{blue}{\underline{0.201}} & 0.273 & 0.206 & 0.274 \\
         & 336 & 0.210 $\pm$ 0.001 & 0.272 $\pm$ 0.001 & \textcolor{blue}{\underline{0.210}} $\pm$ 0.005 & 0.276 $\pm$ 0.002 & 0.215 & \textcolor{blue}{\underline{0.270}} & 0.233 & 0.282 & \textcolor{red}{\textbf{0.209}} & \textcolor{red}{\textbf{0.269}} & 0.221 & 0.279 & 0.220 & 0.275 & 0.215 & 0.276 \\
         & 720 & \textcolor{blue}{\underline{0.210}} $\pm$ 0.000 & 0.276 $\pm$ 0.002 & 0.210 $\pm$ 0.002 & 0.273 $\pm$ 0.004 & 0.218 & \textcolor{blue}{\underline{0.272}} & 0.239 & 0.292 & 0.231 & 0.273 & 0.211 & \textcolor{red}{\textbf{0.266}} & \textcolor{red}{\textbf{0.209}} & 0.276 & 0.224 & 0.286 \\
        \midrule
        \multirow{4}{*}{\raisebox{10pt}{\rotatebox[origin=l]{90}{Traffic}}} & 96 & \textcolor{red}{\textbf{0.384}} $\pm$ 0.002 & \textcolor{red}{\textbf{0.269}} $\pm$ 0.003 & \textcolor{blue}{\underline{0.386}} $\pm$ 0.002 & 0.276 $\pm$ 0.002 & 0.425 & 0.323 & 0.400 & 0.275 & 0.405 & \textcolor{blue}{\underline{0.271}} & 0.426 & 0.311 & 0.433 & 0.297 & 0.573 & 0.377 \\
         & 192 & \textcolor{red}{\textbf{0.403}} $\pm$ 0.003 & \textcolor{red}{\textbf{0.281}} $\pm$ 0.001 & \textcolor{blue}{\underline{0.406}} $\pm$ 0.005 & 0.283 $\pm$ 0.004 & 0.407 & 0.288 & 0.420 & \textcolor{blue}{\underline{0.282}} & 0.446 & 0.282 & 0.435 & 0.307 & 0.455 & 0.299 & 0.770 & 0.460 \\
         & 336 & \textcolor{red}{\textbf{0.419}} $\pm$ 0.008 & 0.286 $\pm$ 0.008 & 0.420 $\pm$ 0.002 & 0.289 $\pm$ 0.003 & \textcolor{blue}{\underline{0.419}} & \textcolor{red}{\textbf{0.283}} & 0.458 & \textcolor{blue}{\underline{0.284}} & 0.444 & 0.289 & 0.538 & 0.392 & 0.553 & 0.350 & 0.866 & 0.500 \\
         & 720 & \textcolor{blue}{\underline{0.459}} $\pm$ 0.003 & 0.309 $\pm$ 0.003 & \textcolor{red}{\textbf{0.449}} $\pm$ 0.002 & \textcolor{blue}{\underline{0.305}} $\pm$ 0.002 & 0.471 & 0.321 & 0.535 & \textcolor{red}{\textbf{0.302}} & 0.503 & 0.308 & 0.522 & 0.343 & 0.566 & 0.324 & 0.561 & 0.337 \\
        \midrule
        1$^{st}$ Count &  & \textcolor{red}{\textbf{19}} & \textcolor{red}{\textbf{15}} & \textcolor{blue}{\underline{5}} & 3 & 2 & \textcolor{blue}{\underline{4}} & 0 & 1 & 1 & 3 & 3 & 4 & 2 & 0 & 0 & 2 \\
        \bottomrule
    \end{tabular}}
\end{table*}

\subsection{Aggregate Significance}
\label{appendix:significance}

With five seeds per setting, a per-cell two-sided Wilcoxon test is uninformative (its smallest attainable $p\approx0.06$), so we assess significance in aggregate using the standard multi-dataset comparison protocol~\cite{demvsar2006statistical}: a Friedman test~\cite{friedman1937use} with a Nemenyi post-hoc~\cite{nemenyi1963distribution} over all 32 (dataset, horizon) settings of Table~\ref{tab:results_full}. The Friedman test rejects rank-equality on both metrics (MSE $p\approx4\times10^{-15}$; MAE $p\approx8\times10^{-8}$). By the Nemenyi post-hoc (critical difference $1.86$ at $\alpha=0.05$), zero-shot \model attains the best mean rank among the eight methods ($1.73$ for MSE, $2.22$ for MAE) and is significantly ahead of all six non-\model baselines on both metrics; the zero-shot and end-to-end variants are statistically indistinguishable from each other. This complements the boundary-adaptivity permutation test of Table~\ref{tab:fixed_rate_pairwise_pvalues} ($p\approx10^{-3}$).

\subsection{Compute and Model Complexity}
\label{appendix:compute-complexity}

Table~\ref{tab:compute_complexity} breaks down trainable, total, main-model, and patcher parameters together with multiply-accumulate operations (MACs), anchored at ETTh1 ($L=336\rightarrow H=720$) with every method on the unified backbone and differing only in the patcher. \model's patcher adds only $\sim$49K parameters on top of the $\sim$2.25M main-model parameters, and its MACs ($5.06$G) are close to the static lower bound ($4.79$G) and well below the external entropy patcher ($7.27$G), which must additionally run a separate pre-trained entropy model. The differentiable relaxations (STE, Gumbel-Softmax) share \model's parameter count but incur slightly higher MACs, confirming that \model's accuracy gains do not come from added model capacity or compute.

\begin{table}[ht!]
\centering
\caption{Compute and model complexity per patching method, anchored at ETTh1 (look-back $336
\rightarrow$ horizon $720$). Every method runs on the unified backbone under the same protocol,
differing only in the patcher. MACs reported in giga-operations.}
\label{tab:compute_complexity}
\small
\renewcommand{\arraystretch}{1.1}
\setlength{\tabcolsep}{5pt}
\begin{tabular}{@{}lccccc@{}}
\toprule
\textbf{Method} & \shortstack{\textbf{Trainable}\\\textbf{Params}} & \shortstack{\textbf{Total}\\\textbf{Params}} & \shortstack{\textbf{Main-Model}\\\textbf{Params}} & \shortstack{\textbf{Patcher}\\\textbf{Params}} & \textbf{MACs (G)} \\
\midrule
\model\ (end-to-end) [Ours] & 2{,}296{,}889 & 2{,}296{,}889 & 2{,}247{,}782 & 49{,}107 & 5.055 \\
STE Relaxation              & 2{,}296{,}889 & 2{,}296{,}889 & 2{,}247{,}782 & 49{,}107 & 5.148 \\
Gumbel-Softmax              & 2{,}296{,}889 & 2{,}296{,}889 & 2{,}247{,}782 & 49{,}107 & 5.148 \\
Static Patching             & 2{,}012{,}638 & 2{,}012{,}638 & 2{,}012{,}622 & 16      & 4.791 \\
Entropy Patching            & 2{,}014{,}670 & 2{,}050{,}670 & 2{,}012{,}622 & 38{,}048 & 7.272 \\
\bottomrule
\end{tabular}
\end{table}

\iffalse
\newpage
\input{checklist}
\fi

\end{document}